\documentclass[review,3p]{elsarticle}

\usepackage[english]{babel}
\usepackage{amsmath}
\usepackage{amssymb}
\usepackage{dsfont}
\usepackage{graphicx}
\usepackage[colorlinks=true, allcolors=blue]{hyperref}
\usepackage{xurl}
\usepackage{enumitem}
\usepackage{makecell}
\usepackage{bbold}
\usepackage{commath}
\usepackage{float}
\usepackage{longtable}

\usepackage[font=footnotesize]{subcaption}  
\usepackage{xparse}                          

\usepackage{standalone}
\usepackage{tikz} 
\usepackage{xfp}
\usetikzlibrary{shapes,arrows}
\usepackage{booktabs}

\usepackage[labelsep=period]{caption}
\usepackage{multirow}
\usepackage{multicol}

\usepackage{lineno}

\usepackage{bm}
\usepackage{mathtools}
\usepackage{comment}
\usepackage{pbox}
\usepackage{xcolor}
\usepackage[table]{xcolor}
\usepackage{tabularx}

\usepackage[ruled,linesnumbered,vlined]{algorithm2e}
\SetKwInput{KwIn}{Input}
\SetKwInput{KwOut}{Output}

\graphicspath{{image/}}

\usepackage{nameref}

\usepackage{titlesec}
\titleformat{\paragraph}[runin]{\normalfont\normalsize\bfseries}{\theparagraph}{1em}{}
\titlespacing*{\paragraph}{0pt}{0pt}{1em}{}

\usepackage{lipsum}

\newif\ifeditornote
\editornotefalse   

\newcommand{\editornotetext}{%
  \ifeditornote
    \makebox[0pt][l]{%
      \parbox[b]{0.65\textwidth}{%
        \footnotesize
        \setlength{\baselineskip}{10pt}%
        The author Junho Song is an editor of this journal. In accordance with policy,
        Junho Song was blinded to the entire peer review process.%
      }%
    }%
  \fi
}

\makeatletter
\def\ps@pprintTitle{%
  \let\@oddhead\@empty
  \let\@evenhead\@empty
  \def\@oddfoot{%
    \editornotetext
    \hfill \thepage
  }%
  \let\@evenfoot\@oddfoot
}
\makeatother

\usepackage{fancyhdr}
\begin{document}


\begin{frontmatter}

\title{Reference-state System Reliability method for\\[-2pt]scalable uncertainty quantification of coherent systems}

\author[1]{Ji-Eun Byun}
\ead{j.byun@imperial.ac.uk}
\address[1]{Department of Civil and Environmental Engineering, Imperial College London, United Kingdom}

\author[2]{Hyeuk Ryu}
\ead{hyeuk.ryu@ga.gov.au}
\address[2]{Geoscience Australia, Australia}

\author[3]{Junho Song\corref{cor1}}
\ead{junhosong@snu.ac.uk}
\address[3]{Department of Civil, Urban, and Environmental Engineering, Seoul National University, South Korea}

\cortext[cor1]{Corresponding author}

\begin{abstract}
Coherent systems are representative of many practical applications, ranging from infrastructure networks to supply chains. Probabilistic evaluation of such systems remains challenging, however, because existing decomposition-based methods scale poorly as the number of components grows. To address this limitation, this study proposes the \textit{Reference-state System Reliability} (RSR) method. Like existing approaches, RSR characterises the boundary between different system states using \textit{reference states} in the component-state space. Where it departs from these methods is in how the state space is explored: rather than using reference states to decompose the space into disjoint hypercubes, RSR uses them to classify Monte Carlo samples, making computational cost significantly less sensitive to the number of reference states. To make this classification efficient, samples and reference states are stored as matrices and compared using batched matrix operations, allowing RSR to exploit the advances in high-throughput matrix computing driven by modern machine learning. We demonstrate that RSR evaluates the system-state probability of a graph with 119 nodes and 295 edges within 10~seconds, highlighting its potential for real-time risk assessment of large-scale systems. We further show that RSR scales to problems involving hundreds of thousands of reference states---well beyond the reach of existing methods---and extends naturally to multi-state systems. Nevertheless, when the number of boundary reference states grows exceedingly large, RSR's convergence slows down, a limitation shared with existing reference-state-based approaches that motivates future research into learning-based representations of system-state boundaries.
\end{abstract}


\end{frontmatter}

\section{Introduction}
Quantifying the reliability of complex systems from the probabilistic performance of their components remains a computationally challenging problem. The core difficulty lies in coupling high-dimensional probabilistic evaluation with system performance simulation---a task whose computational cost grows rapidly with system size and complexity. A large body of work has focused on \textit{coherent} systems, which satisfy the property that the system state never worsens (improves) when component states improve (worsen). Coherent models can represent a wide range of practical systems, including gas distribution networks \citep{ByunSong2021_GMBN}, transportation networks \citep{BRC2026}, supply chains \citep{Chang2025_logistics}, communication networks \citep{Yeh2021_BAT}, and social networks \citep{Zhang2019_MDD_trust}.

This study aims to enable efficient probabilistic evaluation of coherent systems by overcoming the scalability limitations of existing methods. To this end, we propose the \textit{Reference-state System Reliability} (RSR) method, which recasts the system reliability problem as a series of batched matrix operations, enabling it to leverage the advances in high-throughput matrix computing driven by modern machine learning.

This paper is organised as follows. Section~\ref{sec:bf} presents the background, covering key concepts related to coherent systems, a review of previous work---primarily decomposition-based approaches---and their limitations. Section~\ref{sec:rsr} introduces the proposed RSR method. The numerical performance of RSR is investigated in Section~\ref{sec:num_inv}. Finally, Section~\ref{sec:con} provides a summary and suggestions for further research.
 
\section{Background: Reliabiltiy analysis of coherent systems}\label{sec:bf}

\subsection{Problem setting and primary challenge}

Let us consider a system consisting of $N$ components represented by \textit{component random variables} $\bm{X}=(X_1,\ldots,X_N)$ following a joint probability distribution $P(\bm{X})$. Each component variable takes one of $M$ discrete states, so that a \textit{component-state vector} $\bm{x} \in \{0,\ldots,M-1\}^N$ represents an assignment of states to all components. The system state, represented by random variable $S$, is determined by a \textit{system performance function} $\Phi(\bm{x})$: $\{0,\ldots,M-1\}^N \rightarrow \{0,\ldots,M_S-1\}$, where $M_S$ denotes the number of system states. 

Given a target system state $m'\in \{0,\ldots,M_S-1\}$, the probabilities of interest are $P(S \le m')$ and its complement $P(S\ge m'+1)=1 - P(S \le m')$. The most straightforward approach is to enumerate all $M^N$ component-state vectors and classify each according to the system performance function. However, the computational cost would grow exponentially in $N$, making the approach quickly infeasible even for modestly-sized systems. Many methods, including the one proposed in this study, aim to mitigate this exponential growth.

\subsection{Monotonicity of coherent systems and reference states} \label{subsec:bg_coh_sys}

A coherent system is one whose $\Phi(\bm{x})$ is componentwise monotone, i.e.
\begin{equation}
    \bm{x}_1 \preceq \bm{x}_2 \;\; \Rightarrow \;\; \Phi(\bm{x}_1) \leq \Phi(\bm{x}_2),
\end{equation}
where the partial order $\bm{x}_1 = (x_{1,1},\ldots,x_{1,N}) \preceq \bm{x}_2 = (x_{2,1},\ldots,x_{2,N})$ is defined componentwise, i.e.\ $x_{1,n} \leq x_{2,n}$ for all $n = 1,\ldots,N$.
The monotonicity property enables the introduction of \textit{reference states}. For a given threshold $m' \in \{0,\ldots,M_S-2\}$, let us define the \textit{lower} reference set as
\begin{equation}
    \mathcal{L}(m') = \left\{ \bm{x} \in \{0,\ldots,M-1\}^N \;\middle|\; \Phi(\bm{x}) \leq m' \right\},
\end{equation}
i.e., the set of component-state vectors known, through evaluation of $\Phi$, to satisfy $S \leq m'$. We define this set to include only non-dominated component-state vectors, i.e., each state $\bm{x}^* \in \mathcal{L}(m')$ for which there exists no $\bm{x} \in \mathcal{L}(m')$ such that $\bm{x}^* \preceq \bm{x}$. As such, $\bm{x}^*$ provides the most economical characterisation of the condition $S \leq m'$. Analogously, the \textit{upper} reference set is defined as
\begin{equation}
    \mathcal{U}(m') = \left\{ \bm{x} \in \{0,\ldots,M-1\}^N \;\middle|\; \Phi(\bm{x}) \geq m'+1 \right\},
\end{equation}
containing only non-dominated elements, i.e., each $\bm{x}^* \in \mathcal{U}(m')$ satisfies that there exists no $\bm{x} \in \mathcal{U}(m')$ such that $\bm{x}^* \succeq \bm{x}$. Reference sets may be complete or partial, depending on the available information.

The underlying idea of reference states is a recurring theme in the analysis of coherent systems. When component variables and the system variable both take binary state (0 for failure and 1 for survival), reference states for $S \le 0$ and $S \ge 1$---comprising $\mathcal{L}(0)$ and $\mathcal{U}(0)$, respectively---are equivalent to the \textit{cut sets} and \textit{link sets} used in system reliability \citep{Song2009_matrix_dependence}. When components take multiple states and the system corresponds to a single-OD maximum-flow problem with target flow $d$, the two concepts correspond to \textit{$d$-minimal cuts} ($d$-MCs) and \textit{$d$-minimal paths} ($d$-MPs), respectively \citep{Xu2025_dminimal, Zuo2007_minpathvectors}. For generic binary-state systems with multi-state component variables, \citet{BRC2026} termed these \textit{failure rules} and \textit{survival rules}, while in the context of multistate flow networks, they are also referred to as \textit{boundary points} \citep{Chang2024_pathbased}. We generalise these existing concepts into \textit{reference states} applicable to (1) arbitrary coherent system types and (2) both binary- and multi-state components and systems.

\begin{figure}[H]
    \centering

    \begin{subfigure}[t]{0.45\linewidth}
        \includegraphics[width=\linewidth]{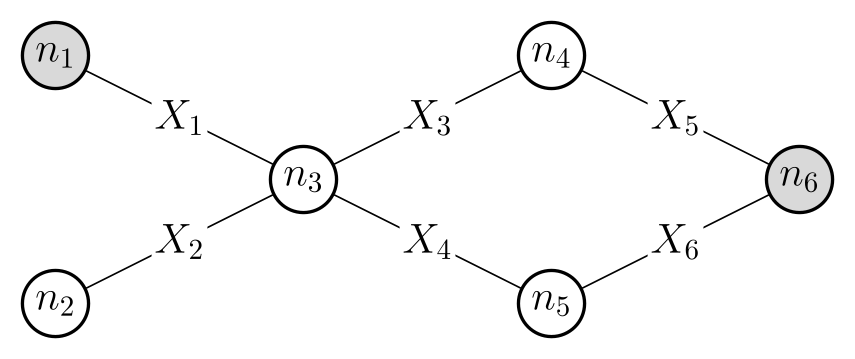}
        \caption{}
        \label{subfig:toy_graph}
    \end{subfigure}
    \hfill
    \begin{subfigure}[t]{0.45\linewidth}
        \includegraphics[width=\linewidth]{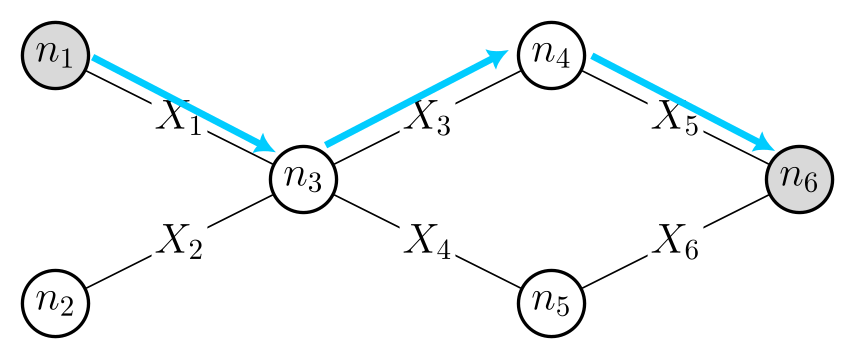}
        \caption{}
        \label{subfig:toy_intact}
    \end{subfigure}

    \vspace{0.5em}

    \begin{subfigure}[t]{0.45\linewidth}
        \includegraphics[width=\linewidth]{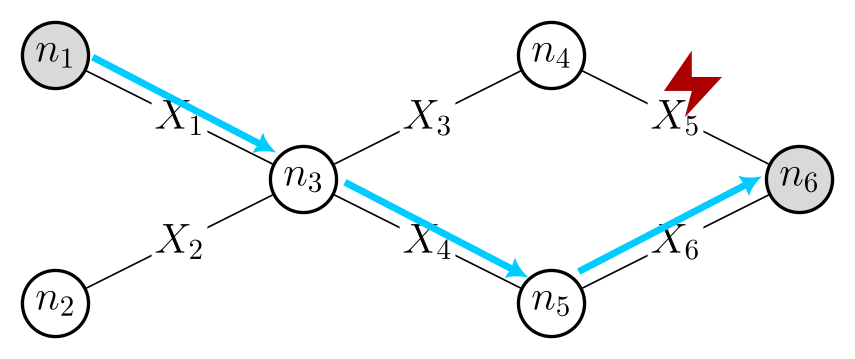}
        \caption{}
        \label{subfig:toy_x5_fail}
    \end{subfigure}
    \hfill
    \begin{subfigure}[t]{0.45\linewidth}
        \includegraphics[width=\linewidth]{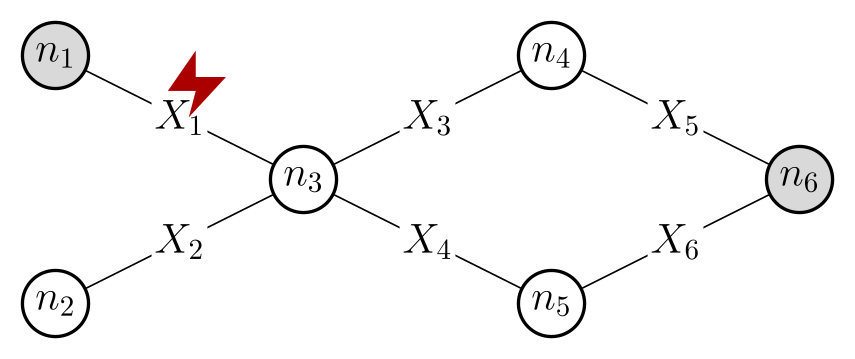}
        \caption{}
        \label{subfig:toy_x1_fail}
    \end{subfigure}
    
\caption{\textbf{Illustrative network system in which the six edges may fail.} The system takes state 0 if nodes 1 and 6 (grey-coloured) are disconnected and 1 otherwise. (a) Original graph. (b) Fully operational network ($\Phi=1$). (c) Failure of edge 5 with $\Phi=1$. (d) Failure of edge 1 leading to $\Phi=0$.}
\label{fig:toy_ex}
\end{figure}

For illustration, Fig.~\ref{subfig:toy_graph} presents an illustrative example of network connectivity, a representative class of coherent system problems. The system consists of six edges that may fail, with their states represented by component random variables $X_1,\ldots,X_6$. The system takes state 0 if nodes 1 and 6 (grey-coloured) are disconnected, and state 1 otherwise. When all components are in state 1, i.e., $\bm{x}_1=(x_1^1,\ldots,x_6^1)$, where $x_n^k$ is shorthand for the assignment $X_n=k$, the system state is $\Phi(\bm{x}_1)=1$, as shown in Fig.~\ref{subfig:toy_intact}. The system state remains unchanged for $\bm{x}_2=(x_1^1,\ldots,x_4^1,x_5^0,x_6^1)$, as shown in Fig.~\ref{subfig:toy_x5_fail}, whereas, with $\bm{x}_3=(x_1^0,x_2^1,\ldots,x_6^1)$, it changes to $\Phi(\bm{x}_3)=0$ as shown in Fig.~\ref{subfig:toy_x1_fail}. 
Among three component-state vectors with known system states, $\bm{x}_1$, $\bm{x}_2$, and $\bm{x}_3$, $\bm{x}_1$ represents redundant information since $\bm{x}_1$ is dominated by $\bm{x}_2$ from $\bm{x}_1 \succeq \bm{x}_2$.

\begin{figure}[H]
    \centering

    \begin{subfigure}[t]{0.32\linewidth}
        \includegraphics[width=\linewidth]{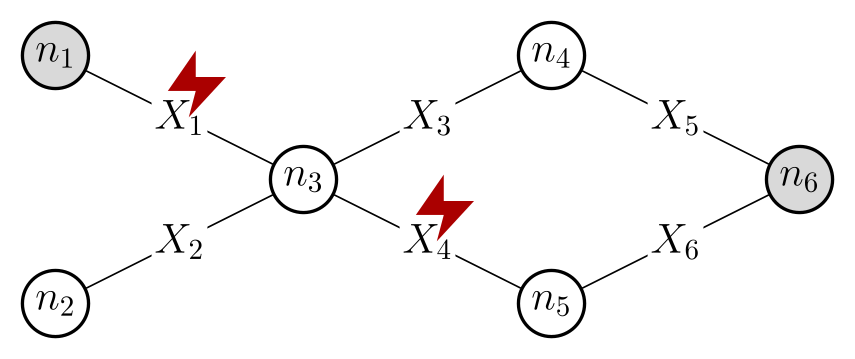}
        \caption{}
        \label{subfig:toy_x1_x4_fail}
    \end{subfigure}
    \hfill
    \begin{subfigure}[t]{0.32\linewidth}
        \includegraphics[width=\linewidth]{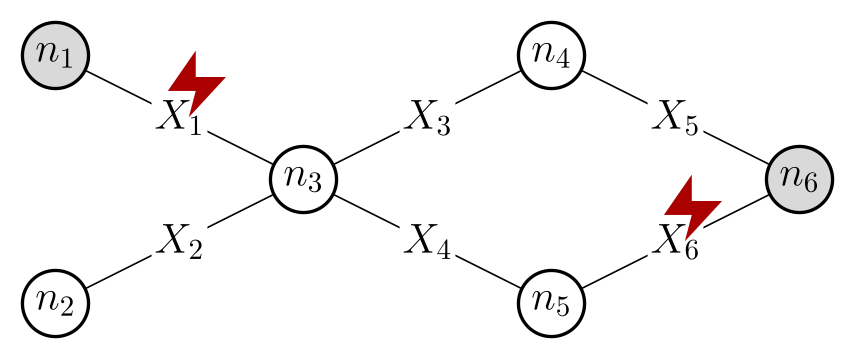}
        \caption{}
        \label{subfig:toy_x1_x6_fail}
    \end{subfigure}
    \hfill
    \begin{subfigure}[t]{0.32\linewidth}
        \includegraphics[width=\linewidth]{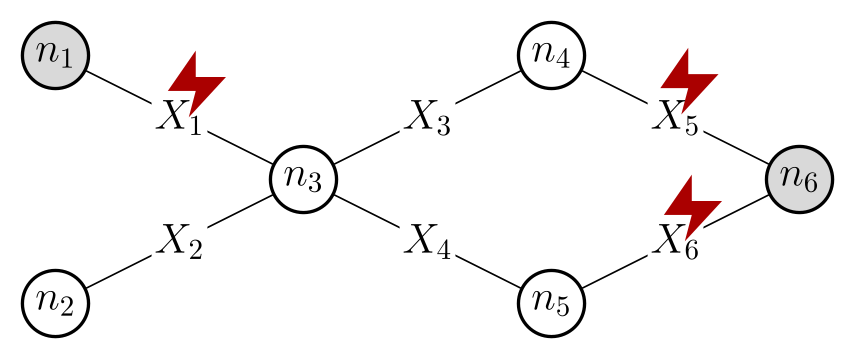}
        \caption{}
        \label{subfig:toy_x1_x5_x6_fail}
    \end{subfigure}
    
    \caption{\textbf{Example component-state vectors dominated by the reference state $\bm{x}_3=(x_1^0,x_2^1,\ldots,x_6^1)$ in Fig.~\ref{subfig:toy_x1_fail}.} Panels show additional component degradations relative to $\bm{x}_3$: (a) edge 4, (b) edge 6, and (c) edges 5 and 6, all of which are dominated by $\bm{x}_3$.}
    \label{fig:toy_domi}
\end{figure}

A reference state enables the classification of all component-state vectors that it dominates. To illustrate this, suppose in the example of Fig.~\ref{fig:toy_ex} that the component random variables are independent with joint distribution $P(\bm{X})=\prod_{n=1}^6 P(X_n)$, and $P(X_n=0)=0.1$ and $P(X_n=1)=0.9$ for all $n$. The reference state $\bm{x}_3$ therefore has probability $P(\bm{x}_3)=0.1 \cdot 0.9^5 = 0.0590$. However, $\bm{x}_3$ dominates all component-state vectors with $X_1=0$, including $(x_1^0,x_2^1,x_3^1,x_4^0,x_5^1,x_6^1)$, $(x_1^0,x_2^1,x_3^1,x_4^1,x_5^1,x_6^0)$, and $(x_1^0,x_2^1,x_3^1,x_4^1,x_5^0,x_6^0)$, whose corresponding network configurations are shown in Figs.~\ref{subfig:toy_x1_x4_fail}, \ref{subfig:toy_x1_x6_fail}, and \ref{subfig:toy_x1_x5_x6_fail}, respectively. Consequently, knowing $\Phi(\bm{x}_3)=0$ allows all such assignments to be classified under $S \le 0$, collectively accounting for a probability of $P(X_1=0)=0.1$, which is approximately 70\% greater than $P(\bm{x}_3)$ itself. The effectiveness of reference states is even more pronounced for large-scale systems.

\subsection{Decomposition-based approaches for coherent systems}\label{subsec:bg_dec}

\subsubsection{Key idea}

\begin{figure}[H]
    \centering
    \begin{subfigure}[t]{0.5\textwidth}
        \includegraphics[width=0.8\textwidth]{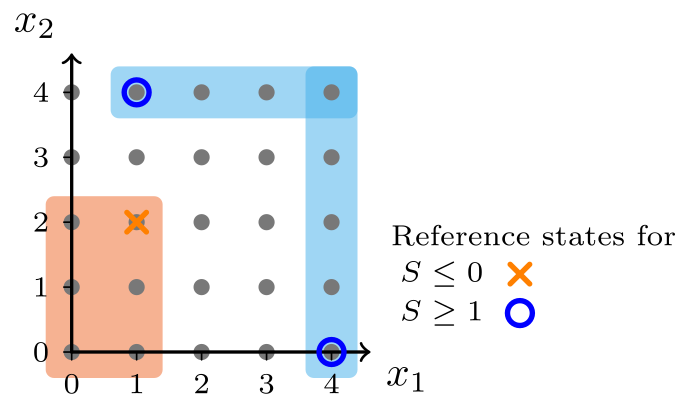}
        \caption{}
        \label{subfig:x1x2_space}
    \end{subfigure}

    \vspace{0.0em}

    \begin{subfigure}[t]{0.32\textwidth}
        \includegraphics[width=0.8\textwidth]{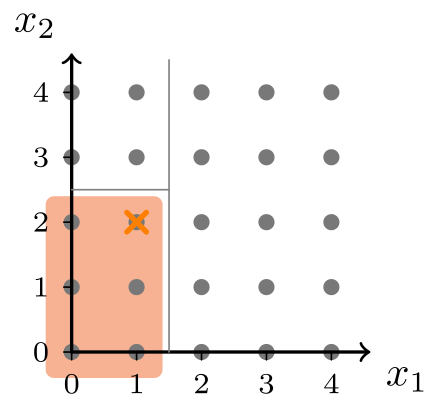}
        \caption{}
        \label{subfig:x1x2_space_ref1}
    \end{subfigure}
    \hfill
    \begin{subfigure}[t]{0.32\textwidth}
        \includegraphics[width=0.8\textwidth]{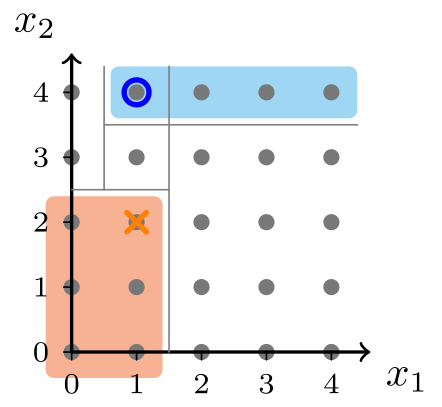}
        \caption{}
        \label{subfig:x1x2_space_ref2}
    \end{subfigure}
    \hfill
    \begin{subfigure}[t]{0.32\textwidth}
        \includegraphics[width=0.8\textwidth]{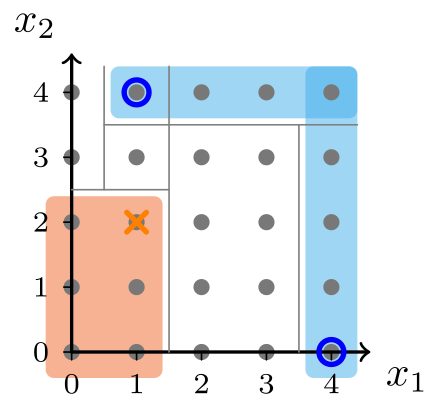}
        \caption{}
        \label{subfig:x1x2_space_ref3}
    \end{subfigure}
    
    \vspace{0.0em}
    
    \begin{subfigure}[b]{0.2\textwidth}
        \centering
        \includegraphics[scale=0.75]{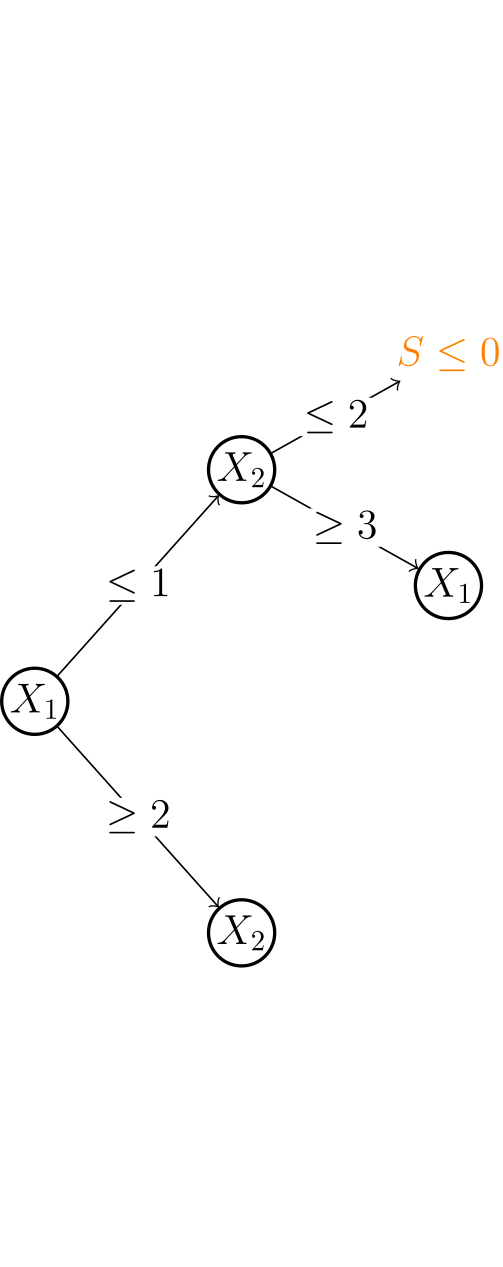}
        \caption{}
        \label{subfig:x1x2_space_dec1}
    \end{subfigure}
    \hfill
    \begin{subfigure}[b]{0.39\textwidth}
        \centering
        \includegraphics[scale=0.75]{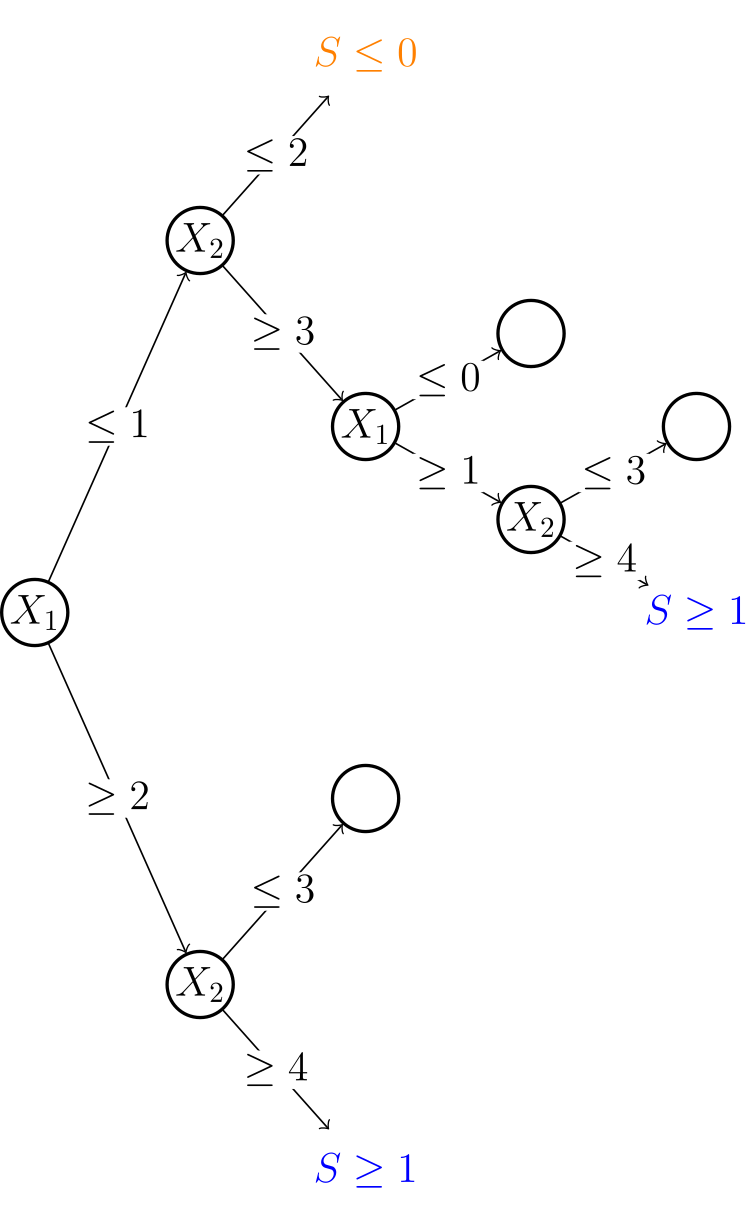}
        \caption{}
        \label{subfig:x1x2_space_dec2}
    \end{subfigure}
    \hfill
    \begin{subfigure}[b]{0.39\textwidth}
        \centering
        \includegraphics[scale=0.75]{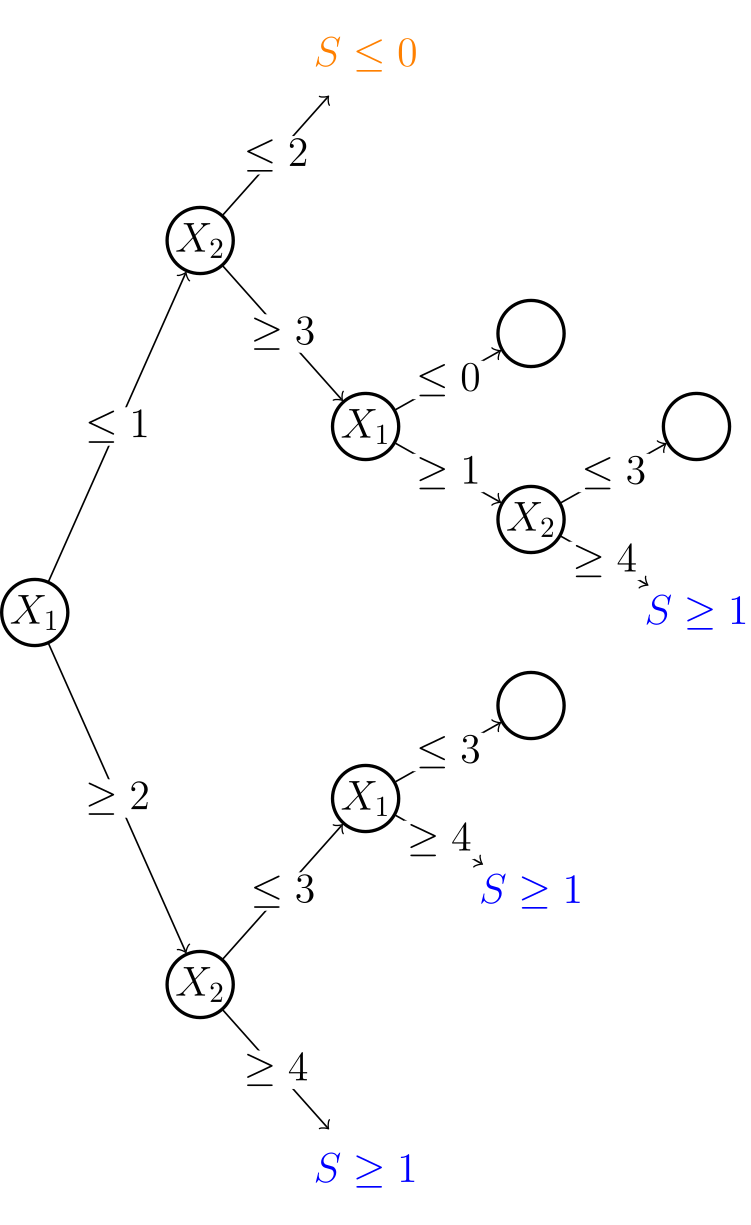}
        \caption{}
        \label{subfig:x1x2_space_dec3}
    \end{subfigure}
    
    \caption{\textbf{Illustrative decomposition process in the component-state space of random variables $X_1$ and $X_2$, each with five states.} Three reference states are assumed to be known, resulting in seven hypercubes.
    (a) The component-state space with the three reference states; shaded regions indicate subspace dominated by references.
    (b)–(d) Step-by-step decomposition of the component-state space as reference states are processed sequentially.
    (e)–(g) Equivalent decomposition steps represented as a branch-and-bound–like process, corresponding to (b), (c), and (d), respectively.}
    \label{fig:x1x2_dec}
\end{figure}

The usefulness of reference states has long been recognised for reliability analysis, particularly leading to the development of various decomposition-based methods. While related work is reviewed in Section~\ref{subsec:lit_rev}, this section summarises the underlying concepts that underpin the literature. 
For illustration, consider the example space of component states in Figure~\ref{subfig:x1x2_space}, consisting of two component random variables $X_1$ and $X_2$, each with five states.\footnote{While it is more common in practice to have $N\gg M$, we use $N=2$ and $M=5$ here for illustrative purposes.} 
The system variable takes two states, 0 and 1, and there are three known reference states: $\mathcal{L}(0) = \{\bm{x}_1^L\}$, where $\bm{x}_1^L = (x_1^1,x_2^2)$ (orange cross in the figure) and $\mathcal{U}(0) = \{\bm{x}_1^U, \bm{x}_2^U\}$ where $\bm{x}_1^U = (x_1^1,x_2^4)$ and $\bm{x}_2^U = (x_1^4,x_2^0)$ (blue circles). The shaded areas represent the subspace dominated by these references.

Given a set of reference states, two computational tasks arise. First, the system probability is calculated as the union probability of the dominated subregions. For example, the blue shaded area in Figure~\ref{subfig:x1x2_space} has a probability evaluated as
\begin{equation}
\begin{aligned}
    P(S\ge 1) &= P \left( \cup_{\bm{x}\in\mathcal{U}(0)} \{ \bm{X}\ge \bm{x} \} \right) \\
    &= P \left(\{\bm{X}\ge \bm{x}_1^U\} \cup \{\bm{X}\ge \bm{x}_2^U\}\right) \\
    &= P \left(\left\{(X_1\ge 1)\cap (X_2\ge 4)\right\} \cup \left\{(X_1\ge 4)\cap (X_2\ge 0)\right\}\right).
\end{aligned}\label{eq:ref_union}
\end{equation}
When evaluated using the inclusion-exclusion principle,\footnote{$P(\cup_{g=1}^G E_g) = \sum_g P(E_g) - \sum_{g<h} P(E_g E_h) + \ldots + (-1)^G P(E_1\ldots E_G)$.} the computational cost grows exponentially with the number of reference states. 

Second, to identify a new reference state, one needs to search the domain not yet covered by known reference states (e.g., the unshaded area in Figure~\ref{subfig:x1x2_space}), defined as
\begin{equation}
\begin{aligned}
    \Omega^u(m')&=\left\{ \bm{x} : \left( \mathcal{L}(m') \right)^\textrm{c} \cap \left( \mathcal{U}(m') \right)^\textrm{c} \right\} \\
    &=\left\{ \bm{x} : \left\{ \cap_{\bm{x}^* \in \mathcal{L}(m') } (\bm{x}\le \bm x^*)^\text{c} \right\} \cap \left\{ \cap_{\bm{x}^* \in \mathcal{U}(m') } (\bm{x}\ge \bm x^*)^\text{c} \right\}  \right\} \\
    &= \left\{ \bm{x} : \left\{ (x_1 \le 1) \cap(x_2\le2) \right\}^\text{c} \cap \left\{ (x_1 \ge 1) \cap(x_2\ge 4) \right\}^\text{c} \cap \left\{ (x_1 \ge 4) \cap(x_2\ge 0) \right\}^\text{c}  \right\}
\end{aligned}\label{eq:omega_u}
\end{equation}
whose computational cost again exponentially increases with the number of known reference states. In complex systems, the number of reference states forming the boundary between two consecutive system states can easily reach thousands or more. Although Eqs.~\eqref{eq:ref_union} and \eqref{eq:omega_u} incur lower costs than the apparent complexity (i.e., the number of component-state vectors $M^N$), direct computation remains infeasible for practical system sizes.

To mitigate this explosive growth, decomposition-based methods partition the component-state space into mutually disjoint subregions, typically in the form of \textit{hypercubes} because of their simple representation and efficient access. A hypercube-shaped event is characterised by a lower bound $\bm{l}\in\{0,\ldots.M-1\}^N$ and an upper bound $\bm{u}\in\{0,\ldots,M-1\}^N$, representing the event $\{\bm{x}: \bm{l} \le \bm{x} \le \bm{u}\}$ (also termed \textit{branches} in \citep{BRC2026}). 
With the reference space $\mathcal{X}$ (either $\mathcal{L}$ or $\mathcal{U}$) decomposed as $\mathcal{X}\rightarrow \mathcal{B}$, the first task then reduces to summing up probabilities over such disjoint events, i.e.,
\begin{equation}
    P\left(\mathcal{X}\right) = P(\mathcal{B}) = P\left(\cup_{b=(\bm{l}, \bm{u}) \in \mathcal{B}} (\bm{l} \le \bm{x} \le \bm{u}) \right) = \sum_{b=(\bm{l}, \bm{u})} P(\bm{l} \le \bm{x} \le \bm{u}).
\end{equation}
Consequently, the computational complexity scales linearly with the number of branches, which, with effective heuristics, can be far fewer than the terms required by inclusion–exclusion over the reference states. The second task also simplifies, as unknown reference states must lie within unclassified hypercubes---those branches whose associated system state is either unknown or not uniquely defined. 

Figures~\ref{subfig:x1x2_space_ref1}-\ref{subfig:x1x2_space_ref3} show the decomposition in the component-state space, while Figures~\ref{subfig:x1x2_space_dec1}-\ref{subfig:x1x2_space_dec3} provide the corresponding branch-and-bound-like representations. Starting from the first reference state $\bm{x}_1^L=(x_1^1,x_2^2)$, the dominated region $\{\bm{x}=(x_1,x_2): (x_1 \le 1) \; \cap \; (x_2 \le 2) \}$ is separated from the remaining space. This produces three hypercubes, two of which remain unclassified. The next reference, $\bm{x}_1^U = (x_1^1,x_2^4)$, further separates the region $\{\bm{x}=(x_1,x_2): (x_1 \ge 1) \; \cap \; (x_2 \ge 4) \}$, decomposing the two unclassified hypercubes into five, three of which remain unclassified. Finally, the reference state $\bm{x}_2^U$ further refines the space, resulting in a total of seven hypercubes. To fully classify the component-state space, the system function is subsequently evaluated at boundary states of the remaining unclassified branches, and the decomposition continues iteratively.

\subsubsection{Previous studies}\label{subsec:lit_rev}

We classify the literature on decomposition-based approaches into three categories: (i) identifying reference states, (ii) computing probabilities using pre-identified reference states, and (iii) performing both tasks in parallel, as summarised in Table~\ref{tab:lit}. The first category has been extensively investigated in network maximum-flow analysis \citep{Xu2025_dminimal, Zhang2025_multistatecuts}, with extensions to multi–origin–destination (OD) connectivity \citep{ChenLin2012_minpaths}. In this context, reference states are commonly referred to as $d$-MCs and $d$-MPs, and research has primarily focused on the efficient identification of all $d$-MCs and $d$-MPs.

The second category assumes that all reference states are available for a given system event. Accordingly, these methods focus on exploiting such prior information to decompose the component-state space into the smallest possible number of subregions. Different approaches have been proposed to this end. For example, branch-and-bound-type algorithms recursively decompose the event space. Applications include single origin–destination (OD) connectivity problems \citep{Zuo2007_minpathvectors, Bai2018_statespace} and multi-OD maximum-flow problems \citep{Chang2025_logistics, LinHuang2014_multisink_accuracy, Niu2025_resourceconstrained}. Decision diagrams have been employed actively (see \citep{Xing2025_decisiondiagrams} for a review), in which the known system logic is represented using fault trees \citep{Zaitseva2025_BDD_uncertainty, Zaitseva2023_MDD_epistemic, Mo2013_BDD_ordering}. We note that some studies proposed exploiting reference states to improve sampling efficiency, thereby leveraging the scalability of sampling-based methods \citep{Chang2024_pathbased, Kozyra2025_simulation, Huang2024_LP_MC, Chang2022_simulation, Niu2025_multicommodity}. In this approach, reference states are used as a computationally cheaper surrogate for system evaluation, replacing the original system function and thereby enabling faster sampling. However, these methods assume the full set of reference states to be available in advance. They therefore inherit the scalability bottleneck of methods developed to identify reference states.

The third category performs reference-state identification and system probability evaluation simultaneously. Methods in this category often assume that exhaustive enumeration of reference states is infeasible, and therefore focus on identifying dominant reference states with priority, while progressively tightening bounds on the system probability using as few reference states as possible. This category can be further classified into two approaches. The first approach recursively partitions the event space, evaluating the system performance function at each branch's bounds to guide further decomposition in a branch-and-bound strategy. Example applications include single origin–destination (OD) connectivity problems \citep{LiHe2002_recursive, LimSong2012_lifeline, LeeSong2021_multiscale}, single-OD maximum-flow problems \citep{JanLai08, Liu2025_configuration}, multiple OD-pair flow problems \citep{DalyAlexopoulos2006_statespace}, and generic system models \citep{BRC2026}. The second approach applies the same interleaved decomposition–evaluation principle but casts the procedure in the framework of decision diagrams. Applications include single-OD connectivity problems \citep{Xing2007_BDD}, with extensions to Binary Additive Trees (BATs) \citep{Yeh2021_BAT, Yeh2022_SA-BAT}, single-OD flow problems \citep{Shrestha2009_MSS_DD, Dong2016_DD_multistateflow}, multiple-OD connectivity problems \citep{WuSun2024_layeredRD, Imai1999_allterminalBDD}, and multi-state $k$-out-of-$N$:G systems \citep{ByunSong2021_GMBN}.

{
\setlength{\tabcolsep}{5pt}
\begin{table}[H]
\centering
\begin{tabularx}{\textwidth}
{>{\raggedright\arraybackslash}X|
 >{\raggedright\arraybackslash}X
 !{\vrule width 1.2pt}
 cccc|c}
        
\Xhline{1.2pt}
\multirow{3}{*}[-20pt]{\makecell[l]{Computational\\[-5pt]task}} &
\multirow{3}{*}[-20pt]{Approach} &
\multicolumn{5}{c}{System types} \\ \cline{3-7}

& &
\multicolumn{4}{c|}{System-specific} &
\multirow{2}{*}[-5pt]{\makecell[l]{System-\\[-5pt]agnostic}} \\ \cline{3-6}

& &
\makecell[l]{1-OD\\[-5pt]conn.} &
\makecell[l]{1-OD\\[-5pt]flows} &
\makecell[l]{Multi-OD\\[-5pt]conn.} &
\makecell[l]{Multi-OD\\[-5pt]flows} &
\\
\Xhline{1.2pt}

\makecell[l]{Reference state\\[-5pt]identification} &
\makecell[l]{d-MCs /\\[-5pt]d-MPs} &
- &
\citep{Xu2025_dminimal, Zhang2025_multistatecuts} &
\citep{ChenLin2012_minpaths} &
- &
- \\ \hline

\multirow{3}{*}[-11pt]{\makecell[l]{Probability\\[-5pt]computation\\[-5pt]over\\[-5pt]reference states}} &
\makecell[l]{Recursive\\[-5pt]decomposition} &
- &
\citep{Zuo2007_minpathvectors, Bai2018_statespace} &
- &
\citep{Chang2025_logistics, LinHuang2014_multisink_accuracy, Niu2025_resourceconstrained} &
- \\ \cline{2-7}

&
\makecell[l]{Decision\\[-5pt]Diagram} &
- & - & - & - &
\makecell[l]{[\citealp{Zaitseva2025_BDD_uncertainty, Zaitseva2023_MDD_epistemic},\\[-5pt]
\citealp{Mo2013_BDD_ordering}]} \\ \cline{2-7}

&
\makecell[l]{References-guided\\[-5pt]sampling} &
- &
\makecell[l]{[\citealp{Chang2024_pathbased, Kozyra2025_simulation},\\[-5pt]
\citealp{Huang2024_LP_MC, Chang2022_simulation}]} &
- &
\citep{Niu2025_multicommodity} &
- \\ \hline

\multirow{4}{*}[-12pt]{\makecell[l]{Simultaneous\\[-5pt]identification\\[-5pt]and\\[-5pt]probability\\[-5pt]computation}} &
\makecell[l]{Recursive\\[-5pt]decomposition} &
\makecell[c]{[\citealp{LiHe2002_recursive, LimSong2012_lifeline},\\[-5pt]\citealp{LeeSong2021_multiscale}]} &
\citep{JanLai08, Liu2025_configuration} &
- &
\citep{DalyAlexopoulos2006_statespace} &
\citep{BRC2026} \\ \cline{2-7}

&
\makecell[l]{Decision\\[-5pt]Diagram\\[-5pt](inc.\ BAT)} &
\makecell[c]{[\citealp{Yeh2021_BAT, Xing2007_BDD},\\[-5pt]
\citealp{Yeh2022_SA-BAT}]} &
\citep{Shrestha2009_MSS_DD, Dong2016_DD_multistateflow} &
\citep{Zhang2019_MDD_trust, WuSun2024_layeredRD, Imai1999_allterminalBDD} &
\citep{ByunSong2021_GMBN} &
- \\ \cline{2-7}

&
\makecell[l]{References-guided\\[-5pt]sampling} &
- & - & - & - &
\cellcolor{gray!15}\makecell[l]{\textbf{Present}\\[-5pt]\textbf{study}} \\
\Xhline{1.2pt}
\end{tabularx}

\caption{\textbf{Selected related studies that exploit reference states for probability evaluation of coherent systems, and the positioning of the present study.} To enable a clear classification, the system categories have been intentionally simplified. The system types considered in the table are not exhaustive; other classes, such as $k$-out-of-$N$ systems and phased-mission systems, are not explicitly represented for simplicity.}
\label{tab:lit}
\end{table}
}

\subsection{Limitations of decomposition-based approaches}\label{subsec:limit}

Despite their significant achievements, decomposition-based approaches face two remaining challenges. 
First, they are typically developed for specific system types, such as network connectivity or maximum flow, which limits their applicability given the diversity of real-world systems. 
While studies such as \citep{Zaitseva2025_BDD_uncertainty, Zaitseva2023_MDD_epistemic, Mo2013_BDD_ordering} are, in principle, applicable to generic coherent systems, they assume the availability of companion fault-tree models, which may not always be available in practice. Recently, initial progress toward relaxing this requirement has been made by the branch-and-bound for reliability analysis of coherent systems (BRC) algorithm \citep{BRC2026}; however, this remains one of the very few attempts reported to date.

\begin{figure}[H]
    \centering

    \begin{subfigure}[t]{0.49\textwidth}
        \centering
        \includegraphics[scale=0.75]{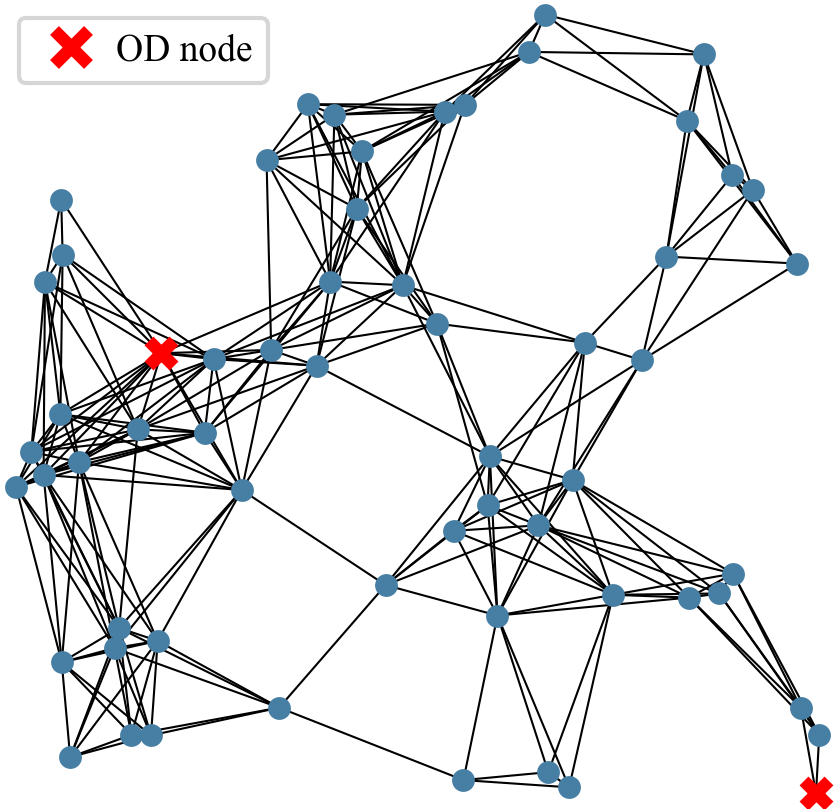}
        \caption{}
        \label{subfig:rg1}
    \end{subfigure}
    \hfill
    \begin{subfigure}[t]{0.49\textwidth}
        \centering
        \includegraphics[scale=0.75]{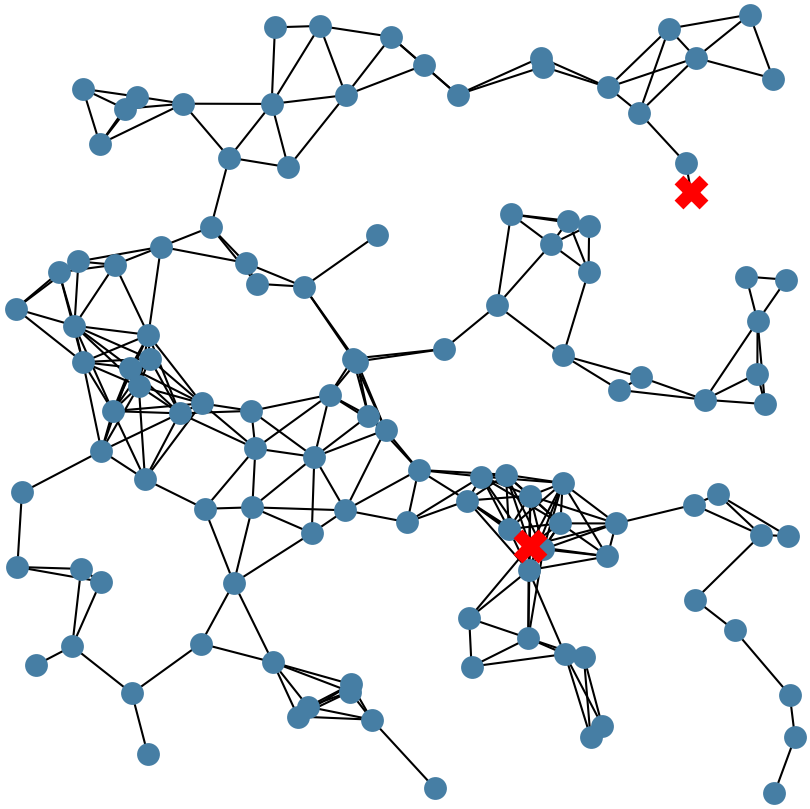}
        \caption{}
        \label{subfig:rg2}
    \end{subfigure}
    
    \vspace{0.2em}
    
    \begin{subfigure}[t]{0.49\textwidth}
        \centering
        \includegraphics[scale=0.7]{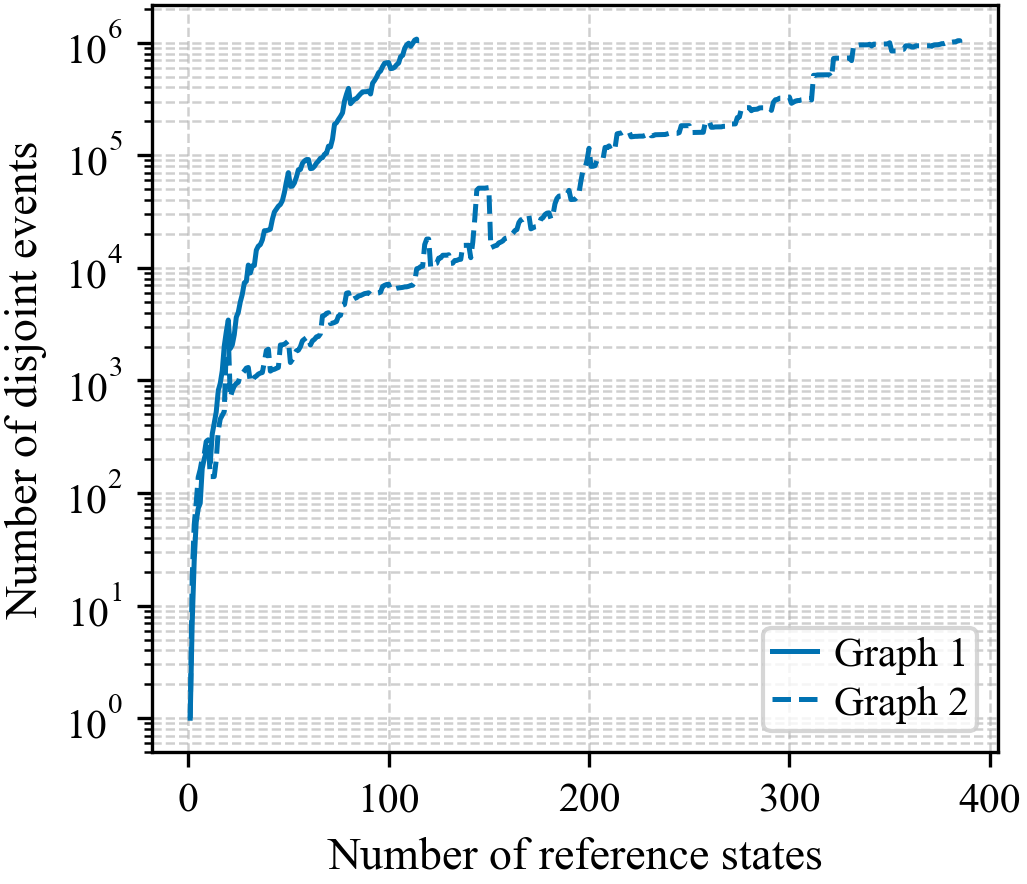}
        \caption{}
        \label{subfig:brc_brs}
    \end{subfigure}
    \hfill
    \begin{subfigure}[t]{0.49\textwidth}
        \centering
        \includegraphics[scale=0.7]{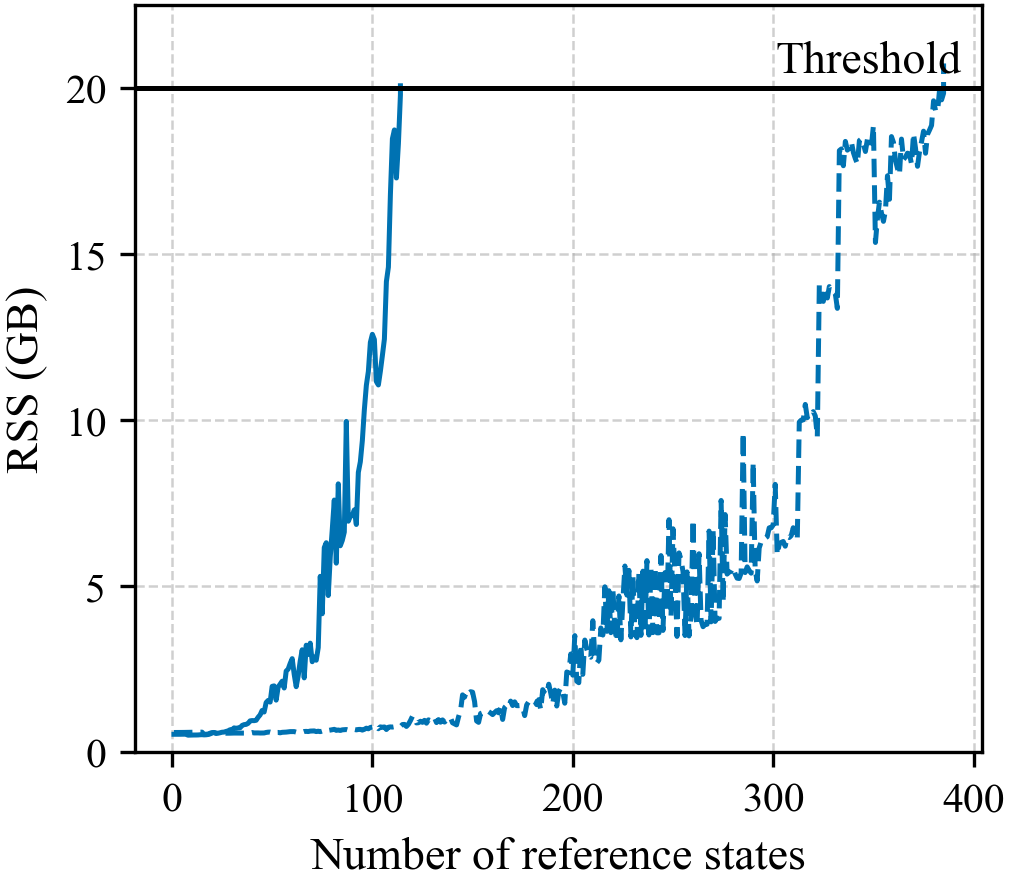}
        \caption{}
        \label{subfig:brc_mem}
    \end{subfigure}

    \vspace{0.2em}

    \begin{subfigure}[t]{0.49\textwidth}
        \centering
        \includegraphics[scale=0.7]{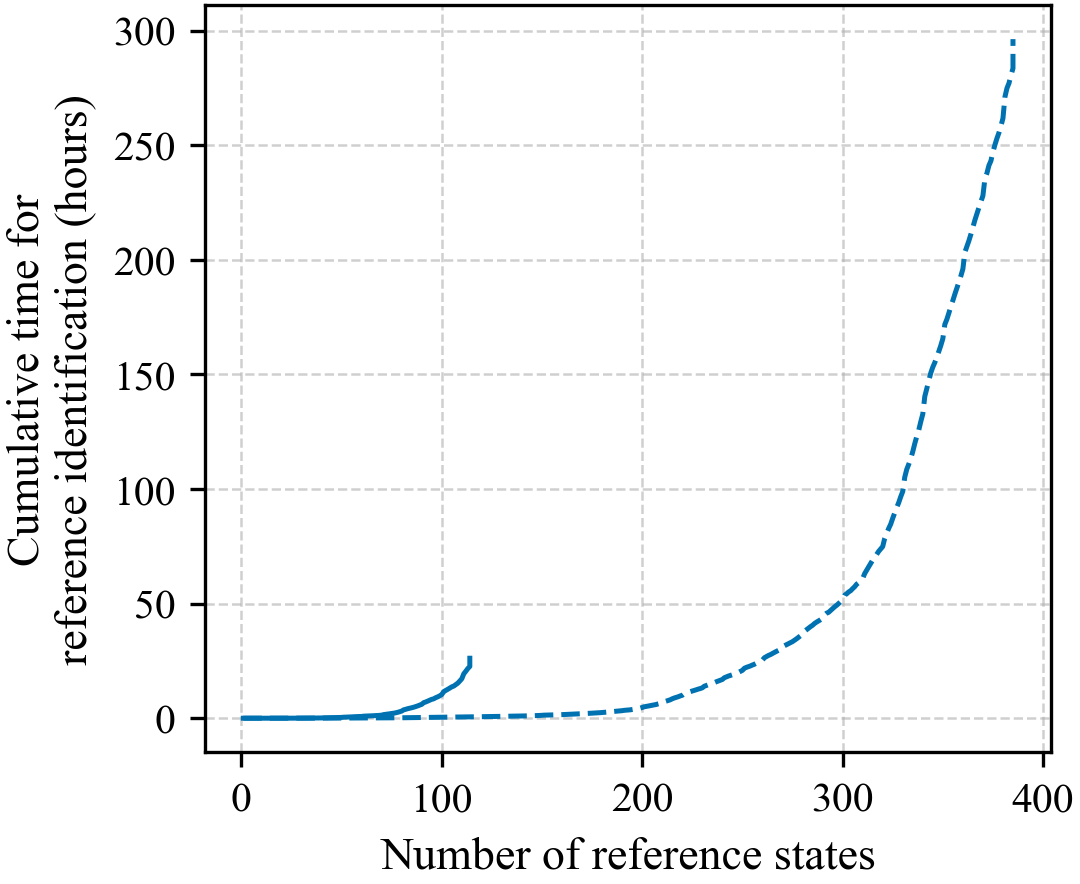}
        \caption{}
        \label{subfig:brc_time}
    \end{subfigure}
    \hfill
    \begin{subfigure}[t]{0.49\textwidth}
        \centering
        \includegraphics[scale=0.7]{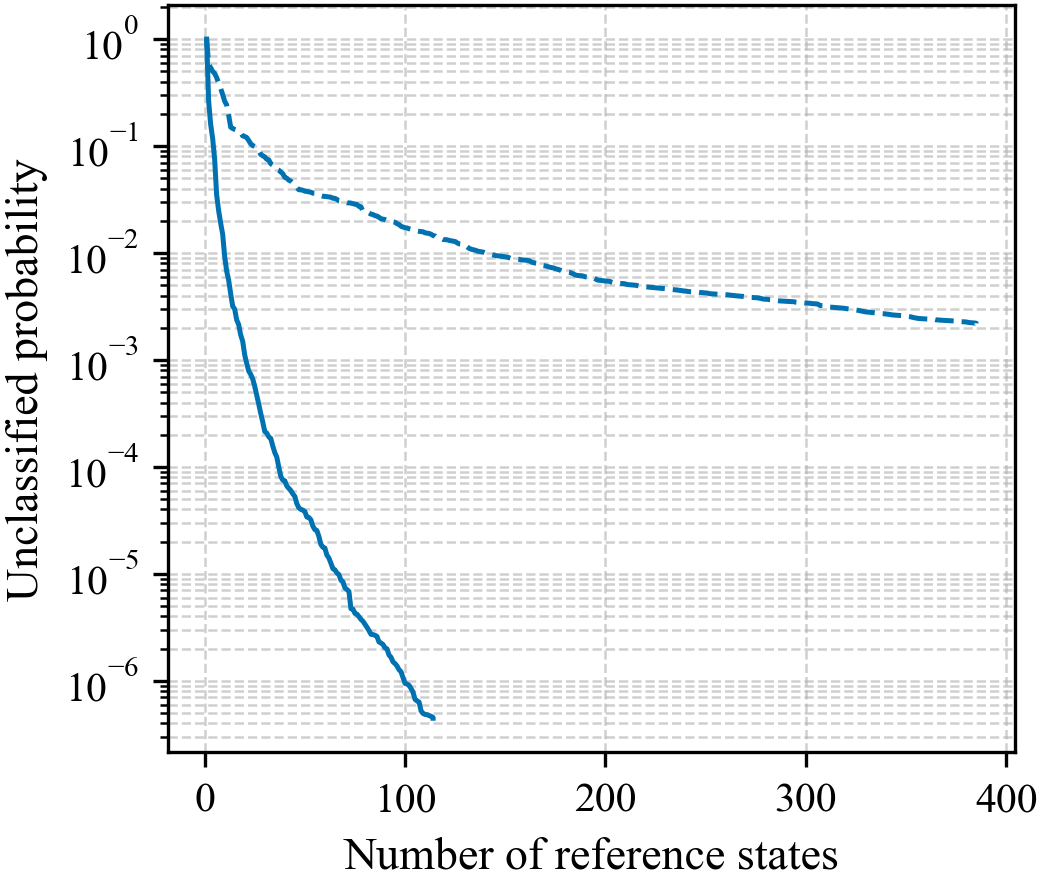}
        \caption{}
        \label{subfig:brc_prob}
    \end{subfigure}
    
    \caption{\textbf{Example random graphs and application of the BRC algorithm \citep{BRC2026}.}
    (a) Graph 1 with 59 nodes and 262 edges and (b) Graph 2 with 119 nodes and 295 edges, where origin and destination nodes are marked by red crosses.
    Panels (c)–(f) show the results of applying the BRC algorithm for connectivity probability estimation. Specifically, (c) number of branches, (d) memory usage, measured by Resident Set Size (RSS), (e) cumulative computational time, and (f) unclassified probability are shown with respect to the number of reference states.
    The connectivity probabilities of the two graphs are evaluated as $0.00250$ and $0.150$, respectively.
    }
    \label{fig:brc}
\end{figure}

Second, decomposition-based approaches remain severely constrained in scalability due to the explosive growth in the number of disjoint events as system size increases. 
To illustrate this issue, we apply the BRC algorithm to connectivity probability estimation.
Two random geometric graphs are generated: one with 59 nodes and 262 edges (Figure~\ref{subfig:rg1}) and the other with 119 nodes and 295 edges, reflecting a larger number of nodes but a lower connectivity density (Figure~\ref{subfig:rg2}). 
We assume that edges are subjected to failure, resulting in 262 and 295 component random variables, respectively. 
All edge failures are assumed to have the same probability of 0.05, i.e., $P(X_n)=0.05$ for $n=1,\ldots,N$, and to be statistically independent.
We compute the single OD connectivity probability, where the origin and destination nodes are indicated by red crosses in the figures.
For consistency, the origin node is chosen as the most connected node, while the destination node is selected as the node topologically most distant from the origin.
This yields probabilities of $S=0$ (i.e., disconnection) of 0.00250 and 0.150 for the two graphs, respectively.

The BRC algorithm is applied to both graphs until the memory usage, measured by Resident Set Size (RSS), reaches 20\,GB. All computations were performed on a desktop computer with an Intel Core i9 processor and 64\,GB RAM.
The results summarised in Figures~\ref{subfig:brc_brs}--\ref{subfig:brc_prob}, illustrate how computational cost escalates rapidly with the number of reference states. Figure~\ref{subfig:brc_brs} shows that the number of disjoint events grows explosively as reference states accumulate, a trend particularly evident on a logarithmic scale. The memory usage and cumulative computation time follow the same pattern (Figures~\ref{subfig:brc_mem} and \ref{subfig:brc_time}), increasing sharply with each additional reference state. Under this growth trend, systems involving thousands of reference states would quickly become infeasible.

\begin{figure}[H]
    \centering
    \includegraphics[scale=0.7]{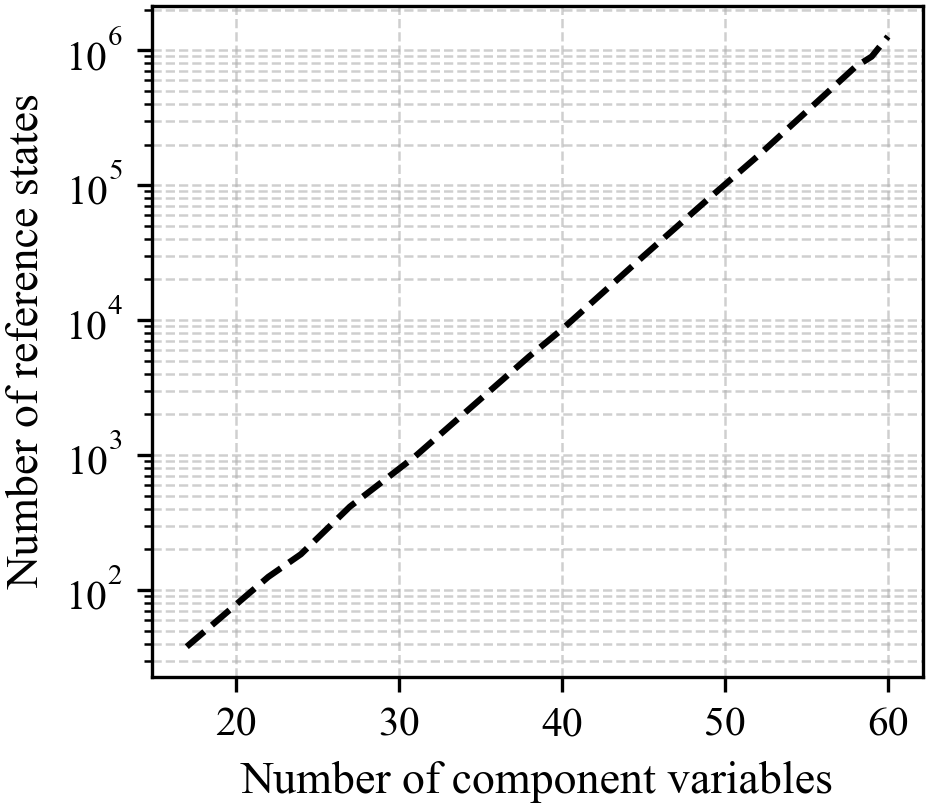}
    \caption{Reported number of reference states in \citep{ChenLin2012_minpaths} (referred to as MPs in that paper) with respect to the number of component variables for maximum-flow probability estimation in a grid topology.}
    \label{fig:n_mps}
\end{figure}

Although the second graph is only moderately larger than the first, the increase in computational cost is far more pronounced, as evident across Figures~\ref{subfig:brc_brs}--\ref{subfig:brc_prob}. A modest increase in system size demands substantially more reference states to achieve the same reduction in uncertainty: in Figure~\ref{subfig:brc_prob}, the unclassified probability for the first graph decreases steadily with additional reference states, while for the second graph it stagnates above 0.001 even beyond 300 reference states.
This sensitivity is well documented, as \citet{ByunStraub2023_ICASP} systematically investigated how the number of reference states grows with the number of components and the definition of system performance. As a concrete example, Figure~\ref{fig:n_mps} reproduces the results of \citet{ChenLin2012_minpaths}, who enumerated reference states (termed MPs in their study) for maximum-flow probability estimation in grid topologies. The figure illustrates that the number of reference states grows exponentially with the number of components, quickly reaching magnitudes that render existing methods infeasible. This motivates the need for a scalable method that can handle large sets of reference states.

\subsection{Alternative approaches for coherent system reliability}

Apart from methods based on reference states, a number of alternative approaches have also been proposed, which are briefly described below for completeness of the literature review. For network reliability, some methods employ the Bayesian network (BN) to directly exploit network topology for probabilistic modelling, which can offer significant improvements in computational efficiency \citep{Lee2025_DGBN_RB, Jiang2025_hierarchical, Zwirglmaier2024_HybridBN}. Yet, they may be limited in the classes of systems they can represent, as BN cannot accommodate directed cycles, which are common in real networks and cannot always be eliminated. Multi-scale approaches, on the other hand, decompose the original system into multiple subsystems consisting of a smaller number of components and combine the probabilities of these subsystems to obtain the overall system probability \citep{LeeSong2021_multiscale, Li2023_bridgeconnectivity, Herrera2025_landmark}. Although such methods are often accompanied by effective approximation formulas to bridge subsystem- and system-level probabilities, their accuracy may degrade when the underlying assumptions, such as Gaussian assumptions on the joint distribution of subsystem events, are violated. Probability density evolution methods provide a computationally efficient alternative by tracing the evolution of system response distributions \citep{Miao2020_PDEM_water, Liu2018_PDEM_gas}. However, their problem-specific derivations may limit their general applicability. More recently, the use of reference states in the context of quantum computing has also been explored \citep{DuenasOsorio2018_quantuminspired}. With continued advances in quantum computing, such approaches may offer new opportunities to accelerate the evaluation of coherent systems. Meanwhile, a range of more numerically oriented methods have been developed, including sampling-based approaches such as sequential Monte Carlo \citep{Vaisman2016_SSMC}, subset simulation \citep{Zuev2015_subset, Lee2025_subset_lifeline}, and importance sampling \citep{Chan2023_BICE, Chan2024_BICE, Kanjilal2023_CEIS}. Another group of methods replaces the original system performance function with surrogate models based on artificial neural networks (ANNs), thereby enabling faster sampling \citep{HuaHuaLin22, HuaHuaLin23, HuangLin2024_LSTM, Shi2024_survivalsignatureML}. 

As these methods do not explicitly exploit reference states—which is a distinctive source of information for coherent systems—their integration with reference-state-based frameworks may offer further performance gains.

\section{Reference-state System Reliability method}\label{sec:rsr}

\subsection{Idea: Replacing decomposition with large-scale sampling}

Motivated by the scalability limitation of decomposition-based methods, we propose the Reference-state System Reliability (RSR) method. Instead of analytically partitioning the event space into disjoint hypercubes, RSR replaces this step with large-scale Monte Carlo sampling, classifying samples directly using available reference states and identifying new reference states from the unclassified samples. System probabilities are then estimated by counting samples under each category of system state. When the exhaustive set of reference states is unavailable, some samples may remain unclassified; these can be resolved by selectively evaluating the system performance function at the corresponding component states.

Implementing this idea requires rapid processing of a large number of samples (e.g., $10^6$). To this end, RSR introduces a matrix-based framework designed to exploit modern computational infrastructure optimised for matrix operations, as widely developed in the context of artificial neural networks. RSR is applicable to generic coherent system types; its positioning relative to existing methods is shown in Table~\ref{tab:lit} (grey cell).

Without loss of generality, we assume in followings that a joint component probability distribution $P(\bm{X})$ is available for generating samples.

\subsection{Matrix-based representation of events}\label{subsec:mat_rep}

A reference state or Monte Carlo sample is represented by a binary matrix of size $N \times M$ through the encoding map
\[
\phi:\{0,\ldots,M-1\}^N \rightarrow \mathbb{B}^{N\times M},
\]
where each row corresponds to a component and each column to a component state. The entry at position $(n,m+1)$ takes the value 1 if state $m$, $m=0,\ldots,M-1$, of component $X_n$, $n=1,\ldots,N$, is included in the corresponding event, and 0 otherwise. Given a target system state $m'\in\{0,1,\ldots,M_S-2\}$, let $\mathcal{L}(m')$ and $\mathcal{U}(m')$ denote the sets of \textit{lower} and \textit{upper} reference states associated with the system events $S\le m'$ and $S\ge m'+1$, respectively (cf., Section~\ref{subsec:bg_coh_sys}). Then, for a lower reference state $\bm{x}_r^{L} \in \mathcal{L}(m')$, the encoded matrix $\mathbf{R}_r^{L}=\phi(\bm{x}_r^{L})$ has entries
\begin{equation}
    r^{L}_{n,m+1,r}= \begin{cases}
        1, & \text{if } m \le \bm{x}_r^{L}\langle X_n \rangle, \\
        0, & \text{otherwise,}
    \end{cases} \quad r=1,\ldots,R^L,
\end{equation}
where $R^L$ denotes the number of lower reference states, and $\bm{x}\langle X_n \rangle$ denotes the state assigned to $X_n$ by the component vector $\bm{x}$. Similarly, for an upper reference state $\bm{x}_r^{U} \in \mathcal{U}(m')$, the encoded matrix $\mathbf{R}_r^{U}=\phi(\bm{x}_r^{U})$ has entries
\begin{equation}
    r^{U}_{n,m+1,r} = \begin{cases}
        1, & \text{if } m \ge \bm{x}_r^{U}\langle X_n \rangle, \\
        0, & \text{otherwise,}
    \end{cases} \quad r=1,\ldots,R^U,
\end{equation}
where $R^U$ denotes the number of upper reference states.
For a Monte Carlo sample $\bm{x}_h \in \mathcal{H}$, the encoded matrix $\mathbf{H}_h=\phi(\bm{x}_h)$ has entries
\begin{equation}
    h_{n,m+1,h} = \begin{cases}
        1, & \text{if } m = \bm{x}_h\langle X_n \rangle, \\
        0, & \text{otherwise,}
    \end{cases} \quad h=1,\ldots,H,
\end{equation}
where $H$ denotes the number of samples. By construction, each row of $\mathbf{H}_h$ contains exactly one nonzero entry. Thus, $\mathbf{H}_h$ is a one-hot matrix representation of the sampled component-state vector $\bm{x}_h$.

The formulation includes the system state $m'$ to support multi-state systems, reflecting the fact that real-world systems inherently operate across multiple states (e.g., a transport network serving multi-commodity, multi-OD demands). For notational brevity, we omit $m'$ when there is no ambiguity (e.g., writing $\mathcal{L}$ in place of $\mathcal{L}(m')$).

\begin{figure}[H]
    \centering
    \includegraphics[width=0.5\textwidth]{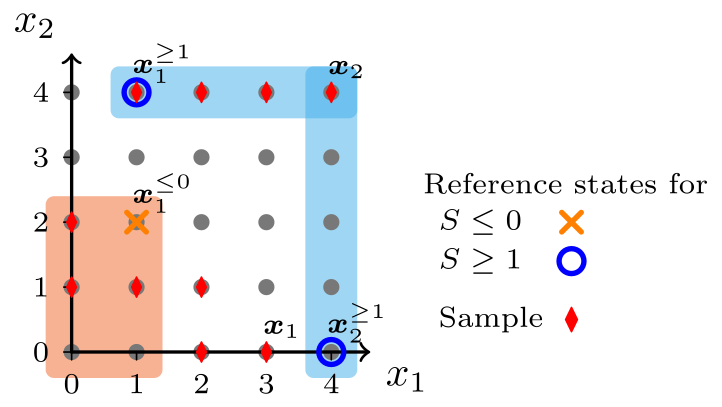}
    \caption{\textbf{Illustration of the $(x_1, x_2)$-state space with reference states and samples.} 
    The orange cross and blue circles represent lower and upper reference states, respectively, 
    while red diamonds indicate sample points in $\bm{x}$-space. 
    The corresponding matrix representations of $\bm{x}_1^{\le 0}$, $\bm{x}_1^{\ge 1}$, $\bm{x}_2^{\ge 1}$, 
    as well as two selected samples $\bm{x}_1$ and $\bm{x}_2$, are given in Eqs.~\eqref{eq:r1_l0}–\eqref{eq:h2}.}
    \label{fig:x1x2_sam}
\end{figure}

For illustration, consider the $\bm{x}$-space (i.e., component-state space) in Figure~\ref{fig:x1x2_sam}, which extends the example in Figure~\ref{fig:x1x2_dec} with ten additional sample points (red diamonds). As before, we assume that three reference states are known. The lower reference state $\bm{x}_1^{L}=(1,2)$ is represented by
\begin{equation}
    \mathbf{R}_1^{L} =
    \begin{bmatrix}
        1 & 1 & 0 & 0 & 0 \\
        1 & 1 & 1 & 0 & 0
    \end{bmatrix},
    \label{eq:r1_l0}
\end{equation}
where rows correspond to $X_1$ and $X_2$, and columns correspond to states $0,1,\ldots,4$. The ones in the matrix represent the orange shaded region.  On the other hand, the upper reference states $\bm{x}_1^{U}=(1,4)$ and $\bm{x}_2^{U}=(4,0)$ are respectively described by
\begin{equation}
    \mathbf{R}_1^{U} =
    \begin{bmatrix}
        0 & 1 & 1 & 1 & 1 \\
        0 & 0 & 0 & 0 & 1
    \end{bmatrix}
    \label{eq:r1_g1}
\end{equation}
and
\begin{equation}
    \mathbf{R}_2^{U} =
    \begin{bmatrix}
        0 & 0 & 0 & 0 & 1 \\
        1 & 1 & 1 & 1 & 1
    \end{bmatrix}.
    \label{eq:r2_g1}
\end{equation}
In Figure~\ref{fig:x1x2_sam}, two samples, $\bm{x}_1=(3,0)$ and $\bm{x}_2=(4,4)$, are highlighted. Their corresponding matrix representations are
\begin{equation}
    \mathbf{H}_1 =
    \begin{bmatrix}
        0 & 0 & 0 & 1 & 0 \\
        1 & 0 & 0 & 0 & 0
    \end{bmatrix},
    \label{eq:h1}
\end{equation}
and
\begin{equation}
    \mathbf{H}_2 =
    \begin{bmatrix}
        0 & 0 & 0 & 0 & 1 \\
        0 & 0 & 0 & 0 & 1
    \end{bmatrix},
    \label{eq:h2}
\end{equation}
which correspond to one-hot encodings of the component states.

\subsection{Reference-state sample classification to calculate system probabilities}\label{subsec:rsr_clas}

\subsubsection{Formulation}

Given the sets of lower and upper reference states, $\mathcal{L}(m')$ and $\mathcal{U}(m')$, RSR performs \textit{reference-state sample classification} on the sample set $\mathcal{H}=\{\bm{x}_h\}_{h=1}^H$. Each sample is classified as falling below a lower reference state, above an upper reference state, or neither, yielding estimates of three probabilities: the lower reference-state probability 
\begin{equation}
    p^{L}(m') = P\left(\bigcup_{\bm{x}'\in\mathcal{L}(m')} (\bm{X}\le \bm{x}')\right),
\end{equation} 
the upper reference-state probability 
\begin{equation}
    p^{U}(m') = P\left(\bigcup_{\bm{x}'\in\mathcal{U}(m')} (\bm{X}\ge \bm{x}')\right),
\end{equation}
and the unclassified probability 
\begin{equation}
    p^{u}(m') = 1 - p^{L}(m') - p^{U}(m').
\end{equation}
The corresponding sample-based estimates are denoted $\hat{p}^{L}(m')$, $\hat{p}^{U}(m')$, and $\hat{p}^{u}(m')$, respectively.
When the reference sets are exhaustive, the unclassified probability vanishes ($p^{u} = 0$), and as $H \to \infty$, the estimates converge to the true probabilities. With both conditions met, $\hat{p}^{L} \to P(S \le m')$ and $\hat{p}^{U} \to P(S \ge m'+1)$.

The classification exploits the matrix-based representation introduced in Section~\ref{subsec:mat_rep}. Since element-wise multiplication corresponds to event intersection,\footnote{This idea follows the matrix-based system reliability (MSR) method \citep{Song2009_matrix_dependence}.} the set-theoretic inclusion between two events $E_1$ and $E_2$,
\begin{equation}
    \text{If } E_1 \cap E_2 = E_1, \text{ then } E_1 \subseteq E_2,
\end{equation}
can be expressed in matrix form as
\begin{equation}
    \text{If } \mathbf{H}\odot \mathbf{R} = \mathbf{H}, \text{ then } E^{\mathbf{H}} \subseteq E^{\mathbf{R}},
\end{equation}
where $E^{\mathbf{H}}$ and $E^{\mathbf{R}}$ denote the events represented by $\mathbf{H}$ and $\mathbf{R}$, respectively, and $\odot$ represents element-wise multiplication. Accordingly, a sample $\bm{x}_h$ is classified into the lower-state domain $\Omega^L=\{\bm{x} \mid \Phi(\bm{x})\le m'\}$ if there exists a lower reference state $\bm{x}_r^{L}\in\mathcal{L}$ such that $\mathbf{H}_h \odot \mathbf{R}_r^{L} = \mathbf{H}_h$. Similarly, it is classified into the upper-state domain $\Omega^U=\{\bm{x} \mid \Phi(\bm{x})\ge m'+1\}$ if there exists an upper reference state $\bm{x}_r^{U}\in\mathcal{U}$ such that $\mathbf{H}_h \odot \mathbf{R}_r^{U} = \mathbf{H}_h$. This induces a partition of the sample $h=1,\ldots,H$ into three disjoint index sets $\mathcal{I}^L$, $\mathcal{I}^U$, and $\mathcal{I}^u$, corresponding to the domains $\Omega^L$, $\Omega^U$, and $\Omega^u$, respectively. The system probabilities are then estimated by counting the samples, i.e.
\begin{subequations}
\begin{align}
    \hat{p}^{L} &= \lvert \mathcal{I}^L\rvert / H, \\
    \hat{p}^{U} &= \lvert \mathcal{I}^U\rvert / H, \\
    \hat{p}^{u} &= \lvert \mathcal{I}^u\rvert / H,
\end{align}
\end{subequations}
where $\lvert \mathcal{I} \rvert$ refers to the cardinality of a set $\mathcal{I}$.
In the illustrative example shown in Figure~\ref{fig:x1x2_sam}, the classification yields $\lvert \mathcal{I}^L\rvert=3$, $\lvert \mathcal{I}^U\rvert=4$, and $\lvert \mathcal{I}^u\rvert=3$, resulting in $\hat{p}^{L}(0)=0.3$, $\hat{p}^{U}(0)=0.4$, and $\hat{p}^{u}(0)=0.3$.

The evaluation of error bounds follows the standard Monte Carlo framework. For an estimated probability $\hat{p}=\lvert \mathcal{I}^L\rvert/H$ or $\lvert \mathcal{I}^U\rvert / H$, the coefficient of variation is given by
\begin{equation}
    \hat{\delta} = \sqrt{\frac{1-\hat{p}}{H \hat{p}}}.
\end{equation}
It is noted that the primary objective of RSR is not to minimise the number of samples, but to enable efficient processing of large sample sets (e.g., $10^6$), so that the sampling error becomes negligible.

\subsubsection{Batched computation}

The subset condition $\mathbf{H} \odot \mathbf{R} = \mathbf{H}$ can be equivalently expressed in vector form. Observing that this condition holds if and only if $\mathbf{H} \odot \bar{\mathbf{R}}$ contains no nonzero entries (where $\bar{\mathbf{R}}$ denotes the element-wise complement of $\mathbf{R}$), a sample $\mathbf{H}$ flattened into a row vector $\bm{h}\in \mathbb{B}^{1 \times NM}$ and a complement reference state $\bar{\mathbf{R}}$ flattened into $\bar{\bm{r}}\in\mathbb{B}^{1 \times NM}$ satisfy
\begin{equation}
    E^{\mathbf{H}} \subseteq E^{\mathbf{R}} \iff \bm{h} \cdot \bar{\bm{r}} = 0.
\end{equation}
This formulation enables simultaneous evaluation across multiple samples and reference states. Specifically, after flattening each sample matrix $\mathbf{H}_h$ into $\bm{h}_h$ and each complement reference-state matrix $\mathbf{R}_r$ into $\bar{\bm{r}}_r$, we can stack all sample vectors into $\mathbf{H}_{\mathrm{flat}} \in \mathbb{B}^{H \times NM}$ and all complement reference-state vectors into $\bar{\mathbf{R}}_{\mathrm{flat}} \in \mathbb{B}^{R \times NM}$, where $R = |\mathcal{L}|$ or $R = |\mathcal{U}|$ depending on the reference set considered. The \emph{violation matrix} is then obtained as
\begin{equation}\label{eq:violation_matrix}
    \mathbf{V} = \mathbf{H}_{\mathrm{flat}}\, 
    \bar{\mathbf{R}}_{\mathrm{flat}}^\top 
    \in \mathbb{Z}_{\ge 0}^{H \times R},
\end{equation}
where $\mathbb{Z}_{\ge 0}$ denotes the set of non-negative integers. Each entry $V_{h,r}$ counts the number of positions at which sample $h$ violates reference state $r$. A sample is therefore classified if at least one reference state is satisfied, i.e.,
\begin{equation}
    \text{sample } h \text{ is classified} \iff 
    \exists\, r \in \{1,\ldots,R\} \;\text{such that}\; V_{h,r} = 0.
\end{equation}

\begin{figure}[H]
    \centering
    \includegraphics[width=0.9\textwidth]{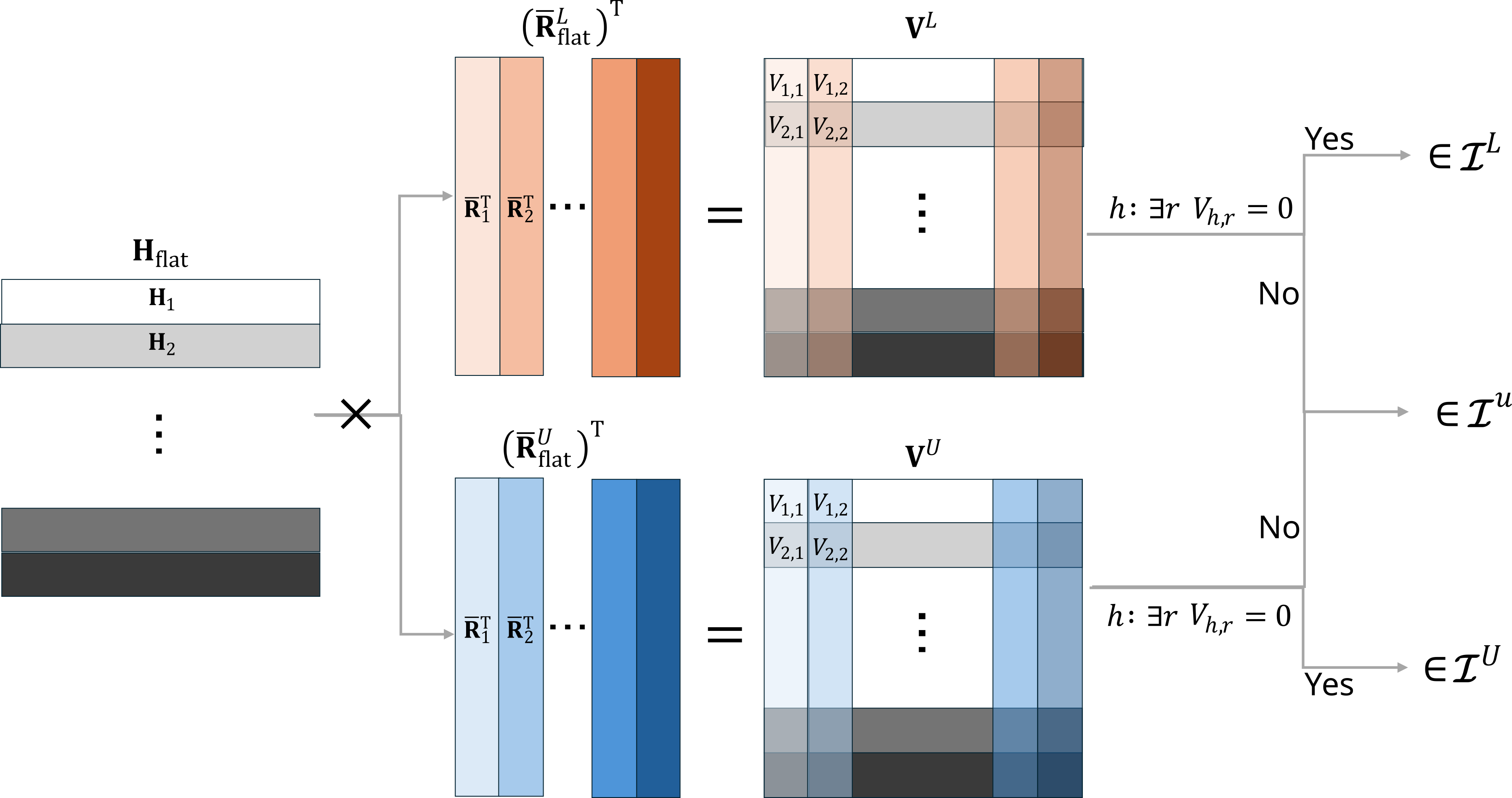}
    \caption{\textbf{Illustration of reference-state sample classification via batched computation.} The flattened sample matrix $\textbf{H}_{\text{flat}}$ is multiplied by the flattened reference-state matrices $\textbf{R}_{\text{flat}}^{L}$ and $\textbf{R}_{\text{flat}}^{U}$.
    If, for a sample $h$, there exists a reference state $r$ such that $V_{h,r}=0$, then $h$ is classified into the corresponding index set for $\mathcal{I}^L$ or $\mathcal{I}^U$; otherwise, it remains unclassified in $\mathcal{I}^u$. The corresponding formal algorithm is given in Algorithm~\ref{alg:rsr_clas}.}

    \label{fig:rsr_clas}
\end{figure}

In practice, RSR computes the violation matrix separately for $\mathcal{L}$ and $\mathcal{U}$, thereby evaluating all reference states in two matrix multiplications. This process is illustrated in Figure~\ref{fig:rsr_clas}, and the formal procedure is summarised in Algorithm~\ref{alg:rsr_clas}. This matrix-based formulation enables efficient parallel implementation, particularly on GPUs.

\begin{algorithm}[htbp]
\caption{Reference-state sample classification (batched computation).}
\label{alg:rsr_clas}

\KwIn{
\\
\quad Sample set $\mathcal{H}=\{\bm{x}_h\}_{h=1}^H$; \\
\quad Lower reference-state set $\mathcal{L}=\{\bm{x}_r^L\}_{r=1}^{R^L}$; \\
\quad Upper reference-state set $\mathcal{U}=\{\bm{x}_r^U\}_{r=1}^{R^U}$.
}

\KwOut{
\\
\quad Lower-classified sample indices $\mathcal{I}^{L}$; \\
\quad Upper-classified sample indices $\mathcal{I}^{U}$; \\
\quad Unclassified sample indices $\mathcal{I}^{u}$.
}
\BlankLine

\tcp{Encode samples and reference states}
$\mathbf{H}_h \leftarrow \phi(\bm{x}_h)$ for $h=1,\ldots,H$\;
$\mathbf{R}_r^L \leftarrow \phi(\bm{x}_r^L)$ for $r=1,\ldots,R^L$\;
$\mathbf{R}_r^U \leftarrow \phi(\bm{x}_r^U)$ for $r=1,\ldots,R^U$\;

\BlankLine
\tcp{Flatten each $N\times M$ matrix into a row vector of length $NM$}
$\mathbf{H}_{\mathrm{flat}} \leftarrow \mathrm{reshape}(\{\mathbf{H}_h\}_{h=1}^H) \in \mathbb{B}^{H \times NM}$\;
$\mathbf{R}_{\mathrm{flat}}^{L} \leftarrow \mathrm{reshape}(\{\mathbf{R}_r^L\}_{r=1}^{R^L}) \in \mathbb{B}^{R^{L} \times NM}$\;
$\mathbf{R}_{\mathrm{flat}}^{U} \leftarrow \mathrm{reshape}(\{\mathbf{R}_r^U\}_{r=1}^{R^U}) \in \mathbb{B}^{R^{U} \times NM}$\;

\BlankLine
\tcp{Negate flattened reference states}
$\bar{\mathbf{R}}_{\mathrm{flat}}^{L} \leftarrow \mathbf{1} - \mathbf{R}_{\mathrm{flat}}^{L} \in \mathbb{B}^{R^{L} \times NM}$\;
$\bar{\mathbf{R}}_{\mathrm{flat}}^{U} \leftarrow \mathbf{1} - \mathbf{R}_{\mathrm{flat}}^{U} \in \mathbb{B}^{R^{U} \times NM}$\;

\BlankLine
\tcp{Dominance screening}
$\mathbf{V}^{L} \leftarrow \mathbf{H}_{\mathrm{flat}} (\bar{\mathbf{R}}_{\mathrm{flat}}^{L})^\top \in \mathbb{Z}_{\ge 0}^{H \times R^{L}}$\;
$\mathbf{V}^{U} \leftarrow \mathbf{H}_{\mathrm{flat}} (\bar{\mathbf{R}}_{\mathrm{flat}}^{U})^\top \in \mathbb{Z}_{\ge 0}^{H \times R^{U}}$\;

\BlankLine
\tcp{Classify: lower takes precedence over upper}
$\mathcal{I}^{L} \leftarrow \{\, h \mid \exists\, r \text{ such that } V^{L}_{h,r} = 0 \,\}$\;
$\mathcal{I}^{U} \leftarrow \{\, h \notin \mathcal{I}^{L} \mid \exists\, r \text{ such that } V^{U}_{h,r} = 0 \,\}$\;
$\mathcal{I}^{u} \leftarrow \{1,\ldots,H\} \setminus (\mathcal{I}^{L} \cup \mathcal{I}^{U})$\;

\BlankLine
\Return $\mathcal{I}^{L}$, $\mathcal{I}^{U}$, $\mathcal{I}^{u}$\;
\end{algorithm}

\subsubsection{Complexity analysis}

In the batched matrix-multiplication formulation of Eq.~\eqref{eq:violation_matrix}, the classification of $H$ samples against $R$ reference states is performed through a single matrix multiplication for each reference-state set, i.e., once for $\mathcal{L}$ and once for $\mathcal{U}$. Each multiplication involves
$\mathbf{H}_{\mathrm{flat}} \in \mathbb{B}^{H \times NM}$ and
$\bar{\mathbf{R}}_{\mathrm{flat}}^\top \in \mathbb{B}^{NM \times R}$,
resulting in a time complexity of $O(HNMR)$.
To control memory usage, the $H$ samples may be processed in chunks of size $H_{\mathrm{chunk}}$. For each chunk, the violation matrix has size $H_{\mathrm{chunk}} \times R$, so the additional working memory per chunk is $O(H_{\mathrm{chunk}}R)$. The persistent memory required to store the flattened samples and reference states is $O(HNM + RNM)$. 

These complexities are summarised in Table~\ref{tab:rsr_clas_comp}. Both time and memory scale linearly with $H$, $R$, $N$, and $M$, while $H_{\mathrm{chunk}}$ provides direct control over the memory--throughput trade-off. This contrasts with the decomposition-based methods discussed in Section~\ref{subsec:limit}, whose complexity grows exponentially with $R$.

{
\setlength{\tabcolsep}{10pt}
\begin{table}[H]
\centering
\begin{tabular}{c|cc}
\Xhline{1.2pt}
Time complexity & \multicolumn{2}{c}{$O(HNMR)$} \\
\Xhline{1.2pt}
\multirow{2}{*}{Memory} & Persistent & Extra per chunk \\
\cline{2-3}
& $O(HNM + RNM)$ & $O(H_{\mathrm{chunk}}R)$ \\
\Xhline{1.2pt}
\end{tabular}
\caption{Time and memory complexity of RSR sample classification
(Algorithm~\ref{alg:rsr_clas}) using the batched matrix-multiplication
formulation. Time complexity scales linearly with the number of
samples $H$, reference states $R$, components $N$, and component
states $M$. Memory per chunk is governed by the tunable parameter
$H_{\mathrm{chunk}}$.}
\label{tab:rsr_clas_comp}
\end{table}
}

\subsection{Componentwise boundary search to identify new reference states}\label{subsec:new_ref}

The sample classification procedure introduced in Section~\ref{subsec:rsr_clas} can be exploited to dynamically identify new reference states. In principle, a new reference state can be obtained by simply selecting a sample from the unclassified samples in $\mathcal{I}^{u}$. However, to preserve computational efficiency, it is crucial that any newly identified reference state be minimal, in the sense that it lies on the boundary of the component-state space separating $\Phi(\bm{x})=m'$ and $\Phi(\bm{x})=m'+1$. If arbitrary samples are indiscriminately added to the reference set, the number of reference states rapidly grows while failing to classify a meaningful portion of samples.

Based on this observation, we propose a \textit{component-wise boundary search} for identifying new boundary reference states. After selecting an unclassified sample, the system function $\Phi(\bm{x})$ is evaluated. If $\Phi(\bm{x}) \le m'$, the components are visited in index order $n=1,\ldots,N$. For each component $n$, its state $\bm{x}\langle X_n \rangle$ is incremented by one unit and $\Phi$ is re-evaluated --- until either the state reaches the maximum state $M$ or the increment causes $\Phi > m'$. If $\Phi > m'$ is encountered, the increment is reverted before proceeding to the next component. After visiting all components, the final $\bm{x}$ represents a boundary reference state for $\Phi(\bm{x}) \le m'$. The case $\Phi(\bm{x}) \ge m'+1$ is handled symmetrically by decrementing component states, leading to a boundary state for $\Phi(\bm{x}) \ge m'+1$. 

\begin{figure}[H]
    \centering
    \begin{subfigure}[t]{0.24\textwidth}
        \includegraphics[width=\textwidth]{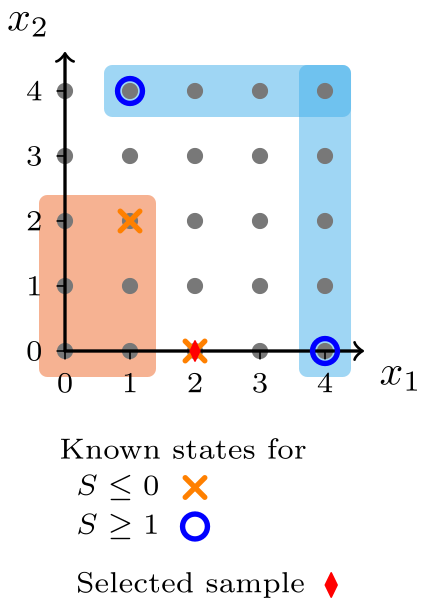}
        \caption{}
        \label{subfig:x1x2_find_ref1}
    \end{subfigure}
    \hfill
    \begin{subfigure}[t]{0.24\textwidth}
        \includegraphics[width=\textwidth]{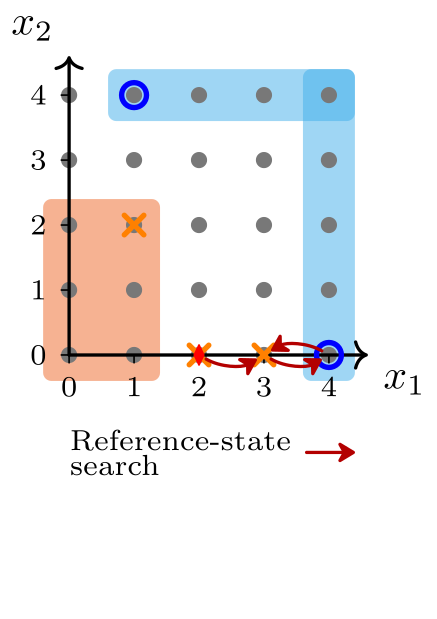}
        \caption{}
        \label{subfig:x1x2_find_ref2}
    \end{subfigure}
    \hfill
    \begin{subfigure}[t]{0.24\textwidth}
        \includegraphics[width=\textwidth]{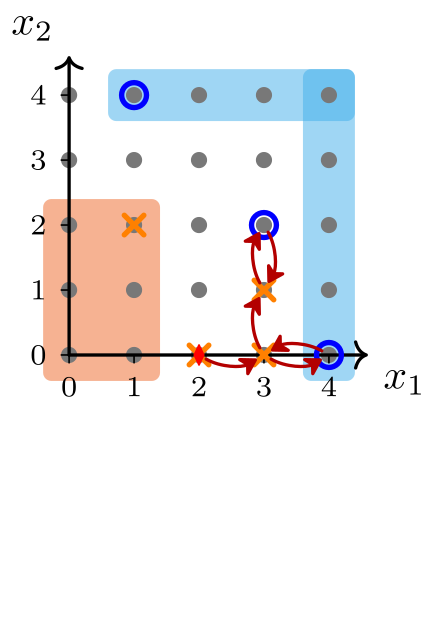}
        \caption{}
        \label{subfig:x1x2_find_ref3}
    \end{subfigure}
    \hfill
    \begin{subfigure}[t]{0.24\textwidth}
        \includegraphics[width=\textwidth]{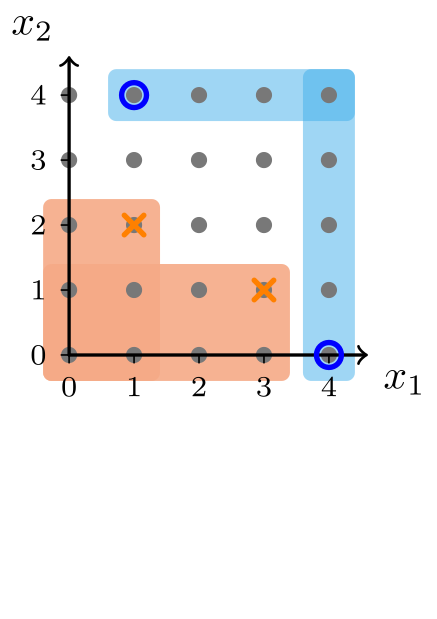}
        \caption{}
        \label{subfig:x1x2_find_ref4}
    \end{subfigure}

    \caption{\textbf{Illustration of component-wise boundary search}, continuing from the sample classification in Figure~\ref{fig:x1x2_sam}.
    (a) A sample $\bm{x}=(2,0)$ is randomly selected among the unclassified samples. 
    Evaluating the system function yields $\Phi(\bm{x})=0$, triggering a search for a new lower reference state associated with $S\le 0$. 
    (b) Search along the $x_1$ direction: the component state is incremented until the state $(4,0)$ is reached, at which point $\Phi=1$. 
    (c) Search along the $x_2$ direction starting from $(3,0)$: the state is incremented until $\Phi((3,2))=1$, yielding a new lower reference state at $(3,1)$. (d) The new lower reference state $(3,1)$ is identified, expanding the known subspace for $S\le0$ (shaded orange).}
    \label{fig:x1x2_find_ref}
\end{figure}

An illustrative example is shown in Figure~\ref{fig:x1x2_find_ref}, continuing from the sample classification in Figure~\ref{fig:x1x2_sam}. In Figure~\ref{subfig:x1x2_find_ref1}, a sample $\bm{x}=(2,0)$ is randomly selected from the unclassified samples and evaluated as $\Phi(\bm{x})=0$, initiating the search for a lower reference state. Figure~\ref{subfig:x1x2_find_ref2} increments the state of $X_1$ until the state $(4,0)$ is reached, where $\Phi((4,0))=1>0$. The search then proceeds along the $X_2$ direction, as shown in Figure~\ref{subfig:x1x2_find_ref3}, starting from $(3,0)$ and terminating at $(3,2)$, where $\Phi((3,2))=1$. Accodingly, in Figure~\ref{subfig:x1x2_find_ref4}, the resulting boundary reference state for $S \le 0$ is identified as $(3,1)$. This expands the known subspace for $S\le 0$ as indicated by the orange shaded area in the figure.

The proposed search procedure requires additional evaluations of the system function, on the order of $O(MN)$ per newly identified reference state. Although the reference-state search introduces additional system-function evaluations, by preventing the explosive growth of non-informative reference states, the strategy leads to a substantial net reduction in the total number of required evaluations. We demonstrate its significance through a numerical example in Section~\ref{subsec:ref_search_ex}.

In practice, to find new reference states, we propose alternating between the sample classification procedure and the identification of new reference states. Since each search is initiated from an unclassified sample, the approach guarantees the discovery of a new reference state at each iteration. Moreover, because samples associated with larger subregions are statistically more likely to be drawn, the strategy naturally prioritises the identification of more influential reference states. The advantage of this stochastic search over the heuristic one employed by decomposition-based methods is demonstrated numerically in Section~\ref{subsec:rsr_brc}.

We note that this strategy is designed for general applicability.
Although more efficient algorithms exist for identifying boundary reference states in specific problem classes--most notably network connectivity \citep{LimSong2012_lifeline, LeeSong2021_multiscale} and maximum flow \citep{Chang2024_pathbased, ChenLin2012_minpaths}--such specialised procedures are not available for a broader class of system functions, such as global connectivity and $k$-out-of-$N$ systems.
Should such a procedure be available for a given system function, this incremental search may be replaced accordingly, thereby reducing the number of system function evaluations.

\subsection{RSR workflow and application to multi-state systems}

\begin{figure}[H]
    \centering
    \includegraphics[width=0.95\linewidth]{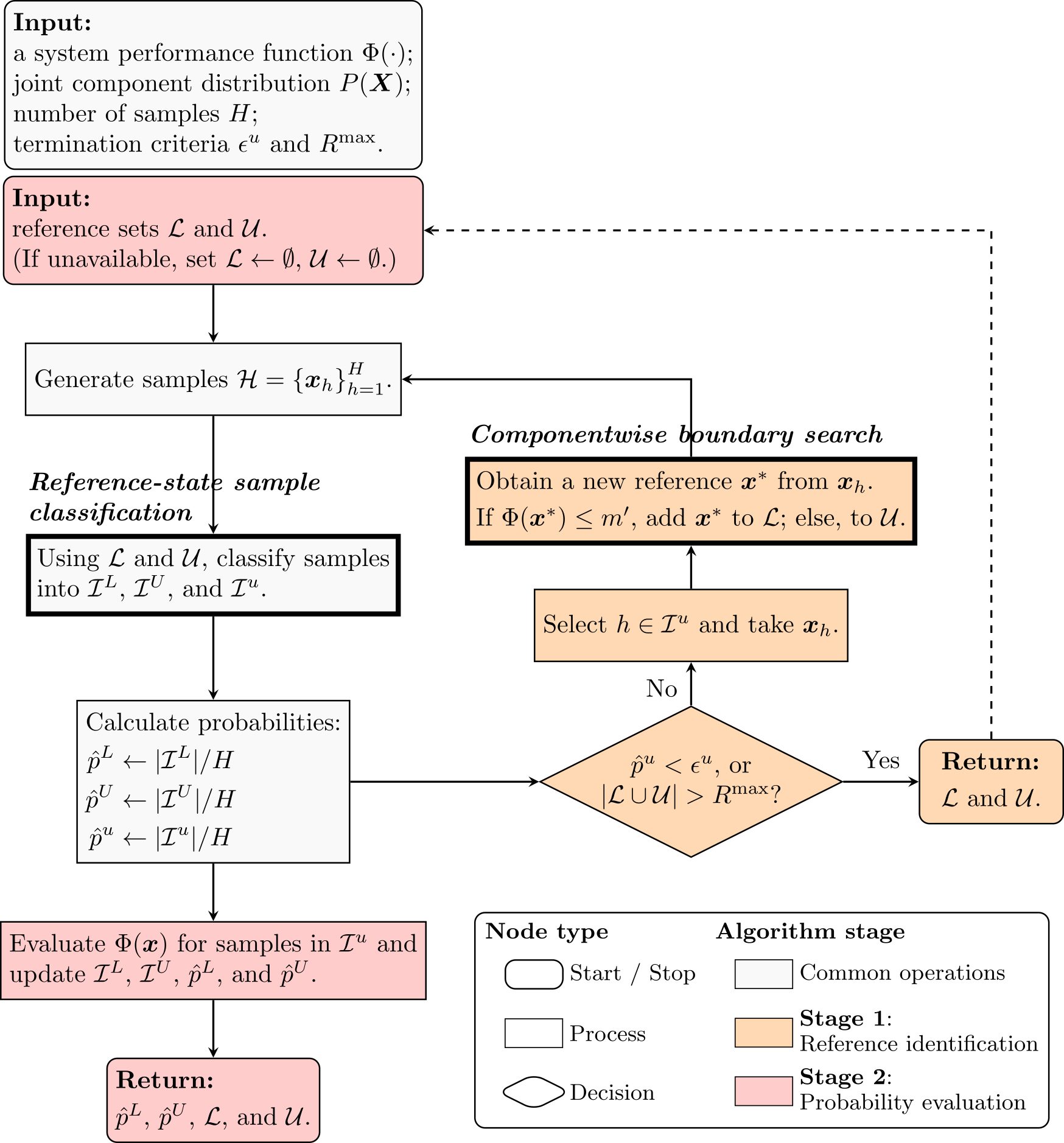}
    \caption{\textbf{RSR algorithm. Stage~1 identifies the boundary reference-state sets $\mathcal{L}$ and $\mathcal{U}$, while Stage~2 evaluates the system probabilities $P(S \le m')$ and $P(S \ge m'+1)$.} 
    Common operations are shown in grey, while Stage~1–specific and Stage~2–specific operations are highlighted in orange and red, respectively. The dashed arrow indicates the use of Stage~1 outcomes as additional inputs to Stage~2. Rounded rectangles, rectangles, and rhombi denote start/stop, process, and decision operations, respectively. Operations incurring the primary computational cost are highlighted with thicker outlines.}
    \label{fig:rsr_alg}
\end{figure}

The RSR workflow comprises two stages, summarised in Figure~\ref{fig:rsr_alg}, where operations for Stage 1 and Stage 2 are indicated by red and orange nodes, respectively, and common operations by grey nodes. Given a target system state $m'$, Stage~1 identifies the sets of boundary reference states, $\mathcal{L}(m')$ and $\mathcal{U}(m')$, while Stage~2 estimates the system probabilities $P(S \le m')$ and $P(S \ge m'+1)$ using the identified reference-state sets. The dominant computational cost arises from sample classification and the identification of new reference states; these operations are highlighted in Figure~\ref{fig:rsr_alg} using thicker outlines.

The two stages share the same inputs, except that the reference-state sets are required only in Stage~2. Both stages begin by generating and classifying $H$ samples. The reference-state sample classification then partitions the samples into three disjoint index sets. In Stage~1, a sample is randomly selected from the unclassified set, and a componentwise boundary search is performed to identify a new boundary reference state. With the updated reference-state sets, the algorithm returns to sample generation and classification. The procedure terminates when either the unclassified probability falls below a user-defined threshold $\epsilon_u$ (e.g., $10^{-5}$) or the number of reference states reaches a prescribed limit $R^{\max}$ (e.g., $10^4$). In Stage~2, by contrast, the system performance function is evaluated for the unclassified samples, from which the system probabilities are estimated.

For systems with more than two states, the workflow is repeated for $m' = 0,1,\ldots,M_S-2$. The probability mass function of the system state can then be obtained as
\begin{equation}
    P(S = m') = P(S \le m') - P(S \le m' - 1),
\end{equation}
or equivalently,
\begin{equation}
    P(S = m') = P(S \ge m') - P(S \ge m' + 1).
\end{equation}

\section{Numerical investigation}\label{sec:num_inv}

\subsection{Comparison with a state-of-the-art decomposition-based method}\label{subsec:rsr_brc}

We compare RSR against BRC \citep{BRC2026}, a state-of-the-art decomposition-based method, on single--OD connectivity problems for the two random graphs introduced in Section~\ref{subsec:limit}. Both algorithms use the same stopping criteria: the unclassified probability falls below $1\times10^{-5}$ or memory usage exceeds 20~GB. All computations are performed on a desktop computer with an Intel Core i9 processor and 64~GB RAM; RSR additionally uses an NVIDIA T400 GPU (4~GB memory) for its matrix operations.\footnote{This is the same computer used in Section~\ref{subsec:limit}} Results are summarised in Table~\ref{tab:rsr_vs_brc}.

Both methods succeed in achieving the unclassified probability threshold of $1\times10^{-5}$ for the single–OD connectivity problem of Graph~1, with 59 nodes and 262 edges. Yet, RSR achieves significant improvements across computational metrics. It requires 25\% fewer reference states (49 vs.\ 66) to achieve the same threshold, implying the advantage of RSR's stochastic search strategy over the heuristic strategy used by decomposition-based approaches. RSR uses only 36\% of the memory (0.81~GB vs.\ 2.24~GB), and completes the reference search over 300 times faster (0.003~hours vs.\ 1.09~hours). 

The performance gap widens dramatically for Graph~2, with 119 nodes and 295 edges. Here, BRC exhausts the 20~GB memory limit after 385 reference states, terminating with an unclassified probability of $2.20\times10^{-3}$---more than two orders of magnitude above the target threshold. RSR, by contrast, reaches the $1\times10^{-5}$ threshold with 288 reference states while consuming less than 0.9~GB of memory and completing the reference search in under 4 minutes, compared to over 12 days for BRC. Once the reference states are obtained, RSR evaluates the system-state probabilities via batched matrix operations in under 10 seconds for both graphs.

\begin{figure}[H]
    \centering
    
    \begin{subfigure}[t]{0.49\textwidth}
        \centering
        \includegraphics[scale=0.72]{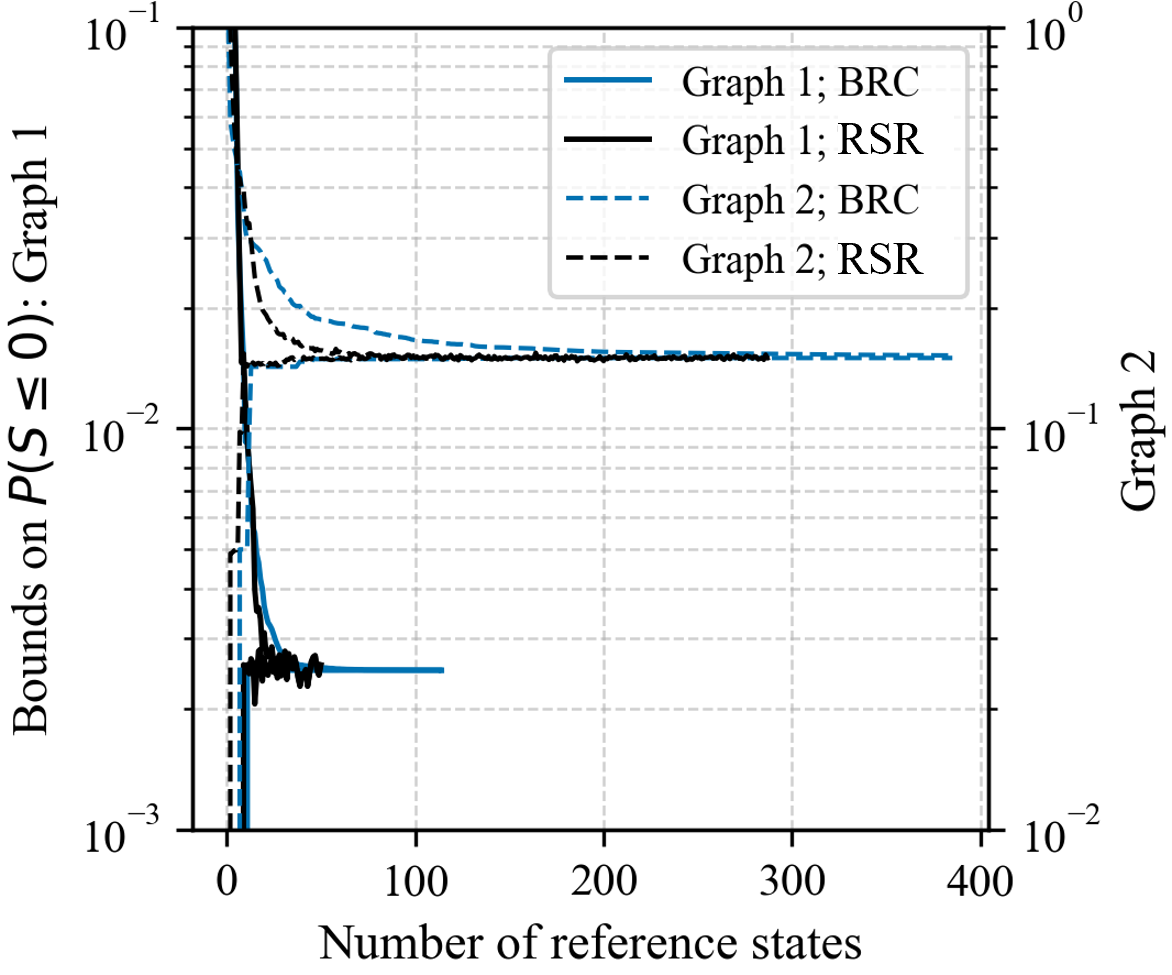}
        \caption{}
        \label{subfig:brc_rsr_bound}
    \end{subfigure}
    \hfill
    \begin{subfigure}[t]{0.49\textwidth}
        \centering
        \includegraphics[scale=0.7]{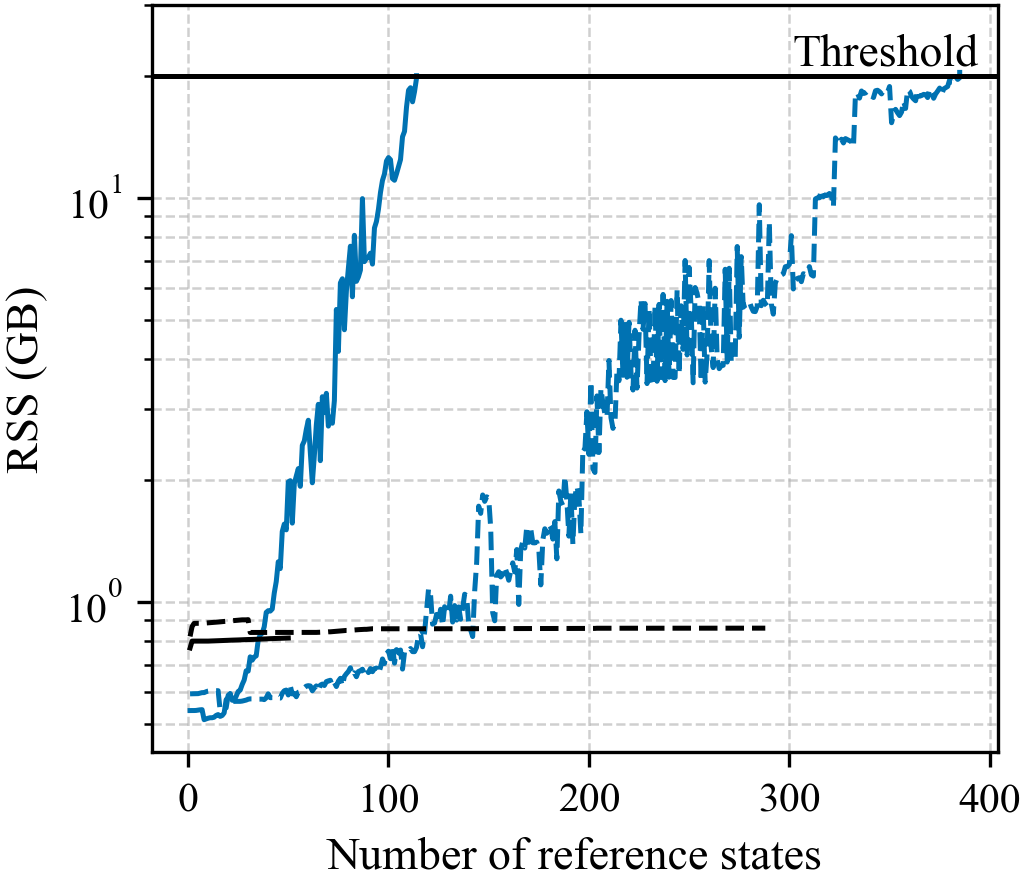}
        \caption{}
        \label{subfig:brc_rsr_mem}
    \end{subfigure}

    \vspace{0.2em}

    \begin{subfigure}[t]{0.49\textwidth}
        \centering
        \includegraphics[scale=0.7]{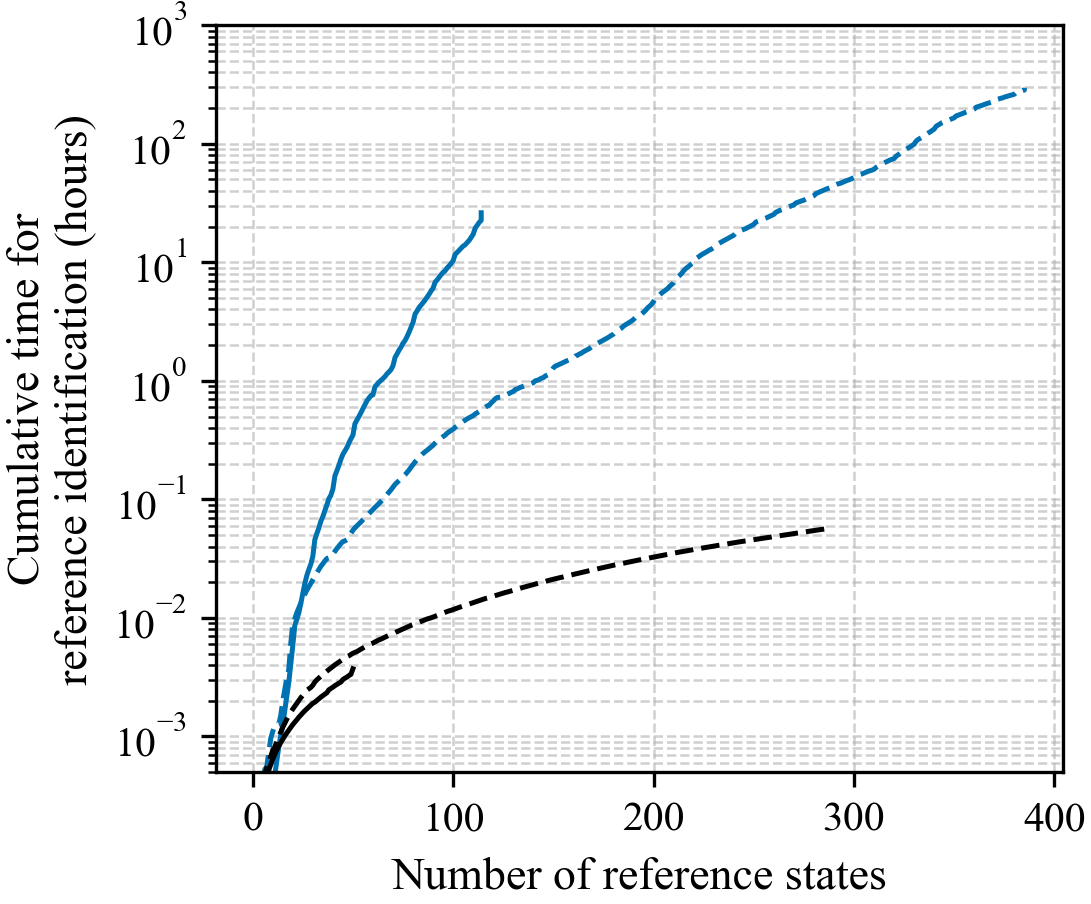}
        \caption{}
        \label{subfig:brc_rsr_time}
    \end{subfigure}
    \hfill
    \begin{subfigure}[t]{0.49\textwidth}
        \centering
        \includegraphics[scale=0.7]{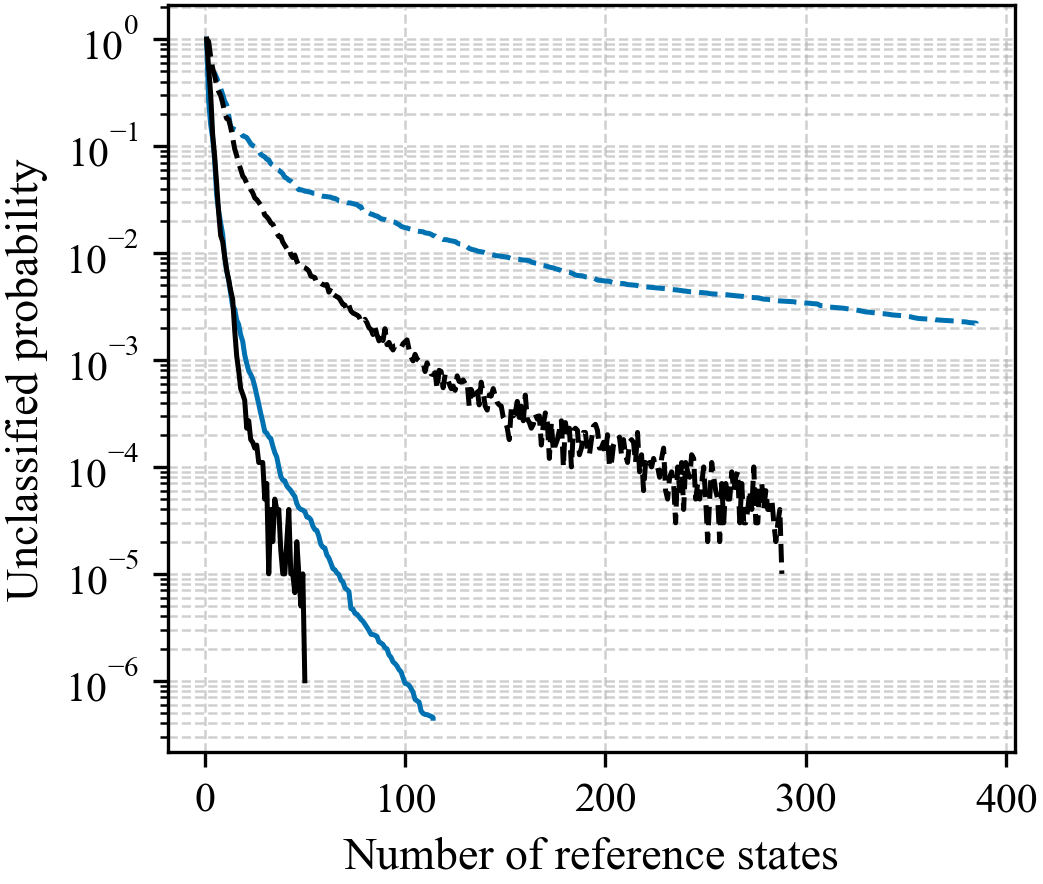}
        \caption{}
        \label{subfig:brc_rsr_prob}
    \end{subfigure}
    
    \caption{\textbf{Comparison of RSR and BRC for the single-OD connectivity on random graphs shown in Figure~\ref{fig:brc}.}
    (a) Bounds on the probability $P(S\le 0)$, (b) memory usage measured by RSS, (c) cumulative computational time, and (d) unclassified probability, all shown as functions of the number of reference states.
    Blue lines represent the BRC data, with panels (b)-(d) showing identical results to those in Figures~\ref{subfig:brc_mem}–\ref{subfig:brc_prob}. Black lines represent the RSR data. 
    Solid lines correspond to Graph~1 and dashed lines to Graph~2.}
    \label{fig:brc_rsr}
\end{figure}

To examine performance in greater detail, Figure~\ref{fig:brc_rsr} compares the progress of reference-state identification for Graph~1, using the same data as Figure~\ref{fig:brc} but with logarithmic y-axes to facilitate comparison. In Figure~\ref{subfig:brc_rsr_bound}, the upper and lower bounds on $P(S \le 0)$ are progressively narrowed for both methods and converge to the same value, as expected. Figure~\ref{subfig:brc_rsr_mem} reveals a notable difference in memory behaviour: while BRC's memory usage grows steadily with the number of reference states, RSR maintains a near-constant and negligible footprint. Figure~\ref{subfig:brc_rsr_time} shows that BRC's cumulative search time increases steeply as more reference states are identified, whereas RSR's grows linearly, indicating a nearly constant cost per additional reference state. Figure~\ref{subfig:brc_rsr_prob} compares the reduction of the unclassified probability, showing that RSR achieves a substantially faster decrease for the same number of reference states. 

\begin{table}[H]
    \centering
    \begin{tabular}{l|c c c|c c c}
    \Xhline{1.2pt}
        \multirow{2}{*}{} 
        & \multicolumn{3}{c|}{Graph 1}  
        & \multicolumn{3}{c}{Graph 2} \\
        \cline{2-7}
        & \makecell[c]{BRC}
        & \makecell[c]{RSR} 
        & \makecell[c]{RSR / BRC}
        & \makecell[c]{BRC}
        & \makecell[c]{RSR} 
        & \makecell[c]{RSR / BRC} \\
    \Xhline{1.2pt} 
         \makecell[l]{No. of references} 
         & 66
         & 49 
         & $74 \ \%$ 
         & 385 & 288 & $75 \ \%$
         \\
         \makecell[l]{Memory (RSS, GB)} 
         & 2.24
         & 0.814 
         & $36 \ \%$ 
         & 20.7
         & 0.861 & $4.3 \ \%$ \\
         \makecell[l]{Reference search\\[-5pt]time (hours)} 
         & 1.09
         & 0.00334 
         & $0.31 \ \%$ 
         & 296 & 0.0571 & $0.019 \ \%$ \\
         \makecell[l]{Unclassified prob.} 
         & $1.00 \cdot 10^{-5}$ & $1.00 \cdot 10^{-5}$ 
         & $100 \ \%$ 
         & $2.20 \cdot 10^{-3}$
         & $1.00 \cdot 10^{-5}$ & $0.45 \ \%$ \\
         \makecell[l]{Prob. evaluation\\[-5pt]time (seconds)} 
         & -
         & 1.33 
         & -
         & -
         & 9.33 & - \\ \hline
         \makecell[l]{$P(S\le0)$} 
         & \multicolumn{3}{c|}{$2.47 \cdot 10^{-3}$} 
         & \multicolumn{3}{c}{$1.50 \cdot 10^{-1}$} \\
    \Xhline{1.2pt}
    \end{tabular}
    \caption{\textbf{Performance comparison of RSR and BRC for single-OD connectivity on the random graphs in Figure~\ref{fig:brc}.} For each graph, the RSR results are reported together with their ratio relative to the corresponding BRC baseline. Both algorithms are executed until the unclassified probability fell below $1\times10^{-5}$ or the memory usage exceeded 20~GB. Probability evaluation time and $P(S\le0)$ are reported for RSR only, as BRC does not support batched probability evaluation. RSR achieves significant reductions across all performance metrics.}
    \label{tab:rsr_vs_brc}
\end{table}

\subsection{Effects of system performance definition on computational complexity}\label{subsec:rol_sys_per}

Having established RSR's advantages over BRC for single-OD connectivity, we now consider a more challenging problem: global connectivity. Here, the system is assigned state~0 if any node becomes disconnected from the rest of the network, and state~1 otherwise. Because each edge plays a relatively balanced role in maintaining global connectivity, the number of boundary reference states required to achieve the same unclassified probability grows substantially compared to the single-OD case. This renders the BRC algorithm infeasible for global connectivity, as the results in Section~\ref{subsec:rsr_brc} show that it struggles to handle more than 300 reference states.

Table~\ref{tab:global_vs_singleod} summarises the RSR results for global connectivity, with percentage changes reported relative to the single-OD baselines in Table~\ref{tab:rsr_vs_brc}. For Graph~1, RSR identifies 8{,}680 reference states---177 times more than for single-OD connectivity---to reduce the unclassified probability below $1\times10^{-5}$. Despite this increase, the problem remains tractable: memory usage rises only modestly to 0.910~GB, and reference-state identification completes in 17.2~hours. For Graph~2, however, the global connectivity problem exposes practical limits. After identifying 10{,}000 reference states over 247~hours of computation, the unclassified probability remains at $6.40\times10^{-2}$, far above the target threshold. This highlights an inherent limitation of explicitly enumerating reference states for large-scale systems and motivates the need for further methodological advances.

\begin{figure}[H]
    \centering
    \begin{subfigure}[t]{0.31\textwidth}
        \centering
        \includegraphics[width=\linewidth]{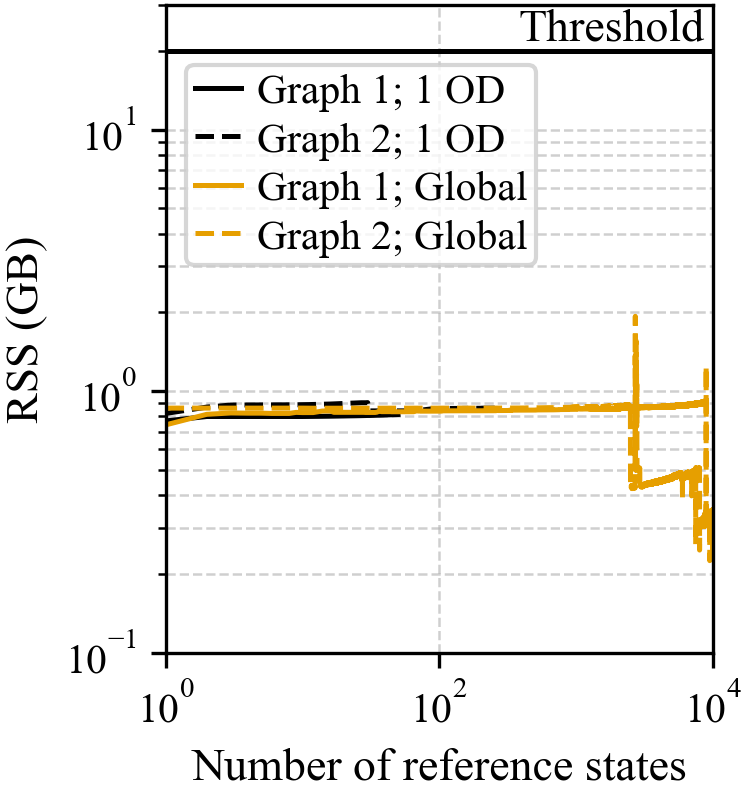}
        \caption{}
        \label{subfig:rsr_mem}
    \end{subfigure}
    \hfill
    \begin{subfigure}[t]{0.31\textwidth}
        \centering
        \includegraphics[width=\linewidth]{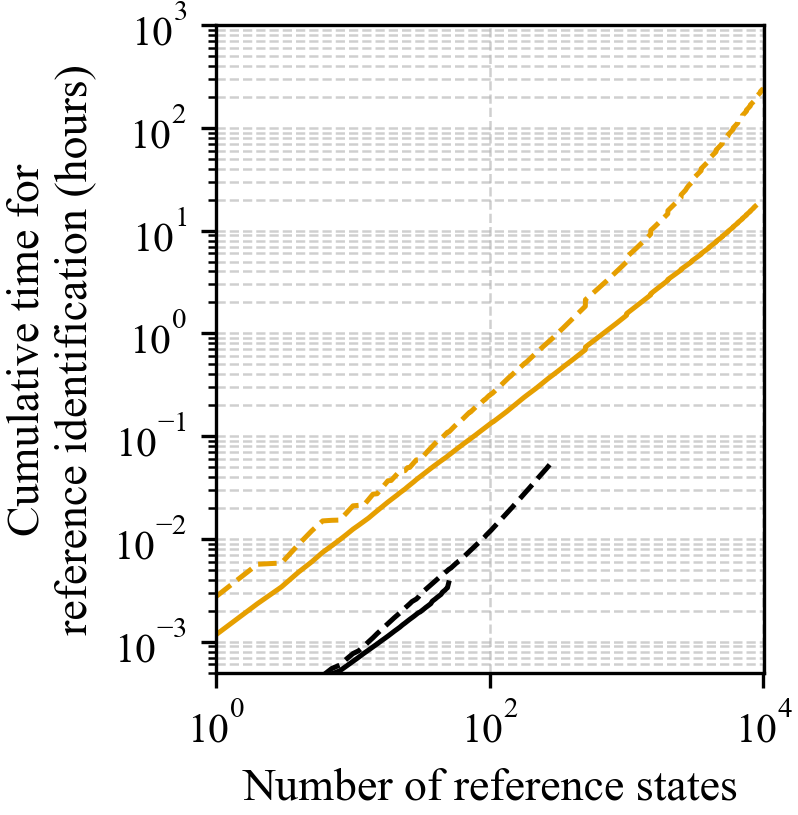}
        \caption{}
        \label{subfig:rsr_time}
    \end{subfigure}
    \hfill   
    \begin{subfigure}[t]{0.31\textwidth}
        \centering
        \includegraphics[width=\linewidth]{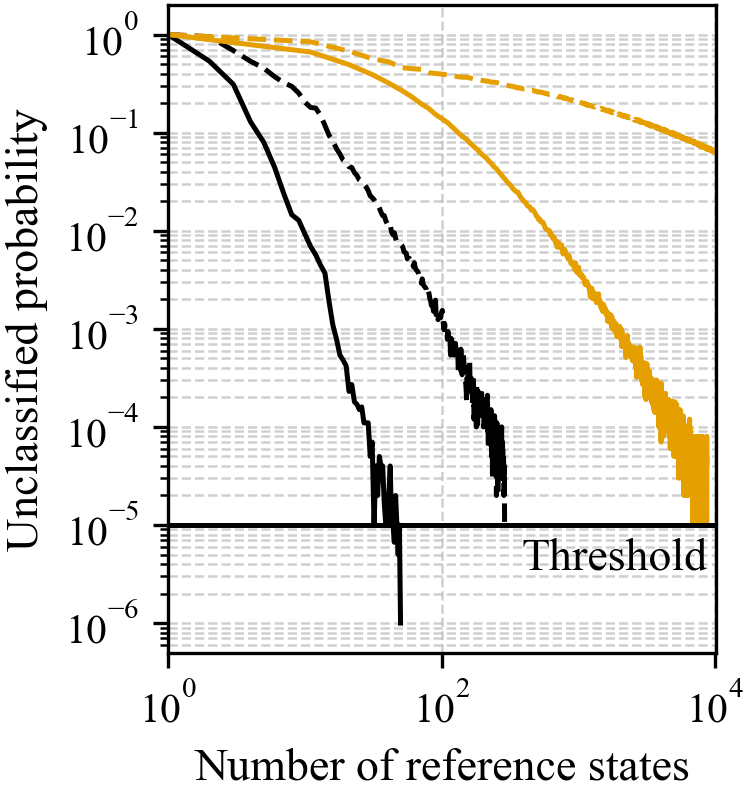}
        \caption{}
        \label{subfig:rsr_prob}
    \end{subfigure}
    
    \caption{\textbf{RSR implementation for single-OD and global connectivity on the random graphs shown in Figure~\ref{fig:brc}.}
    (a) Memory usage measured by RSS, (b) cumulative computational time, and (c) unclassified probability, all shown as functions of the number of reference states.
    Black lines represent single-OD connectivity and are identical to the data shown in Figures~\ref{subfig:brc_rsr_mem}--\ref{subfig:brc_rsr_prob}, with the x-axes shown on a log scale to facilitate comparison.
    Orange lines represent global connectivity.
    Solid lines correspond to Graph~1 and dashed lines to Graph~2.}
    \label{fig:rsr_res}
\end{figure}

Figure~\ref{fig:rsr_res} provides further insight into these scaling behaviours. Memory usage (Figure~\ref{subfig:rsr_mem}) remains negligible even at 10{,}000 reference states, and cumulative identification time (Figure~\ref{subfig:rsr_time}) grows at a nearly constant rate, confirming that RSR's per-iteration cost does not degrade with problem size. The key bottleneck is the rate at which the unclassified probability decreases (Figure~\ref{subfig:rsr_prob}), which is markedly slower for Graph~2 than for Graph~1 under global connectivity. Finally, we examine the probability evaluation stage (i.e., Stage~2). For single-OD connectivity, this step is near-instantaneous: 1.33~seconds for Graph~1 and 9.33~seconds for Graph~2. Under global connectivity, Graph~1 remains efficient at 29.5~seconds, but Graph~2 requires 1.65~hours. Of this, approximately 20~minutes are spent on sample classification and the remaining 1.31~hours on evaluating system states for unclassified samples. 

\begin{table}[H]
    \centering
    \begin{tabular}{l|c c|c c}
    \Xhline{1.2pt}
        \multirow{2}{*}{} 
        & \multicolumn{2}{c|}{Graph 1}  
        & \multicolumn{2}{c}{Graph 2} \\
        \cline{2-5}
        & \makecell[c]{Global}
        & \makecell[c]{Global / Single-OD}
        & \makecell[c]{Global}
        & \makecell[c]{Global / Single-OD} \\
    \Xhline{1.2pt} 
         \makecell[l]{No. of reference states} 
         & 8{,}680 
         & $17{,}714 \ \%$ 
         & 10{,}000 & $3{,}472 \ \%$
         \\
         \makecell[l]{Memory (RSS, GB)} 
         & 0.910 
         & $112 \ \%$ 
         & 0.346 & $40 \ \%$ \\
         \makecell[l]{Reference search\\[-5pt]time (hours)} 
         & 17.2 
         & $515{,}000 \ \%$ 
         & 247 & $432{,}600 \ \%$ \\
         \makecell[l]{Unclassified prob.} 
         & $1.00 \cdot 10^{-5}$ 
         & $100 \ \%$ 
         & $6.40 \cdot 10^{-2}$ & $640{,}000 \ \%$ \\
         \makecell[l]{Prob. evaluation\\[-5pt]time (seconds)} 
         & 29.50 
         & $2{,}218 \ \%$
         & \makecell[c]{5{,}946\\[-5pt]{\small ($=1{,}221 + 4{,}725$)}} & $63{,}752 \ \%$ \\ \hline
         \makecell[l]{$P(S\le0)$} 
         & \multicolumn{2}{c|}{$2.56 \cdot 10^{-3}$} 
         & \multicolumn{2}{c}{$5.41 \cdot 10^{-1}$} \\
    \Xhline{1.2pt}
    \end{tabular}
    \caption{\textbf{RSR performance for global connectivity on the random graphs in Figure~\ref{fig:brc}.} Percentage changes are relative to the single-OD results in Table~\ref{tab:rsr_vs_brc}. The termination criteria are as described in Table~\ref{tab:rsr_vs_brc}, with an additional limit of 10,000 reference states. For Graph~2, the probability evaluation time consists of 1{,}221~seconds for sample classification and 4{,}725~seconds for evaluating the system state of unclassified samples.}
    \label{tab:global_vs_singleod}
\end{table}

\subsection{Performance of the component-wise boundary search} \label{subsec:ref_search_ex}

\begin{figure}[H]
    \centering
    \begin{subfigure}[t]{0.31\textwidth}
        \centering
        \includegraphics[width=\linewidth]{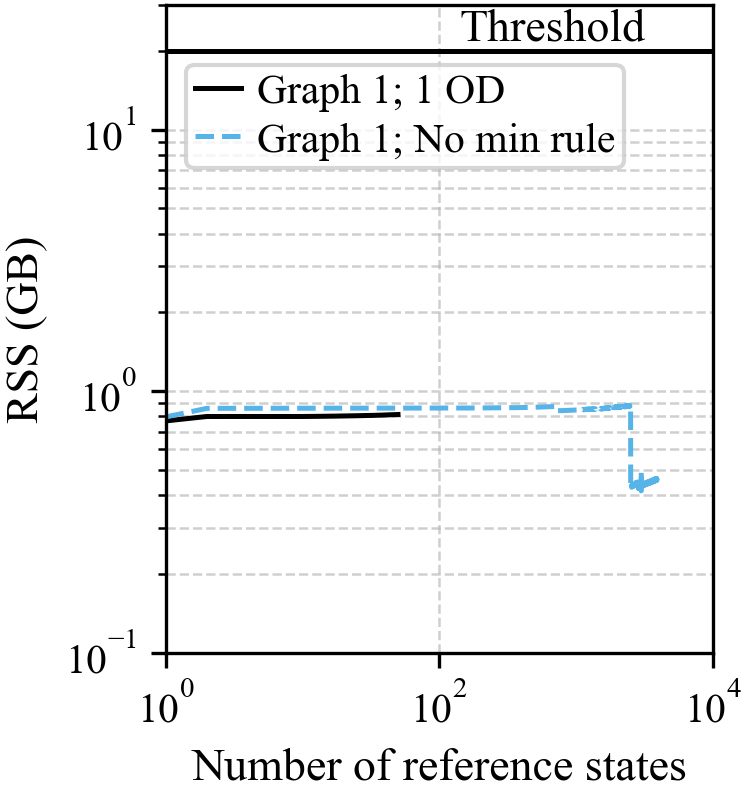}
        \caption{}
        \label{subfig:rsr_mem_nomin}
    \end{subfigure}
    \hfill
    \begin{subfigure}[t]{0.31\textwidth}
        \centering
        \includegraphics[width=\linewidth]{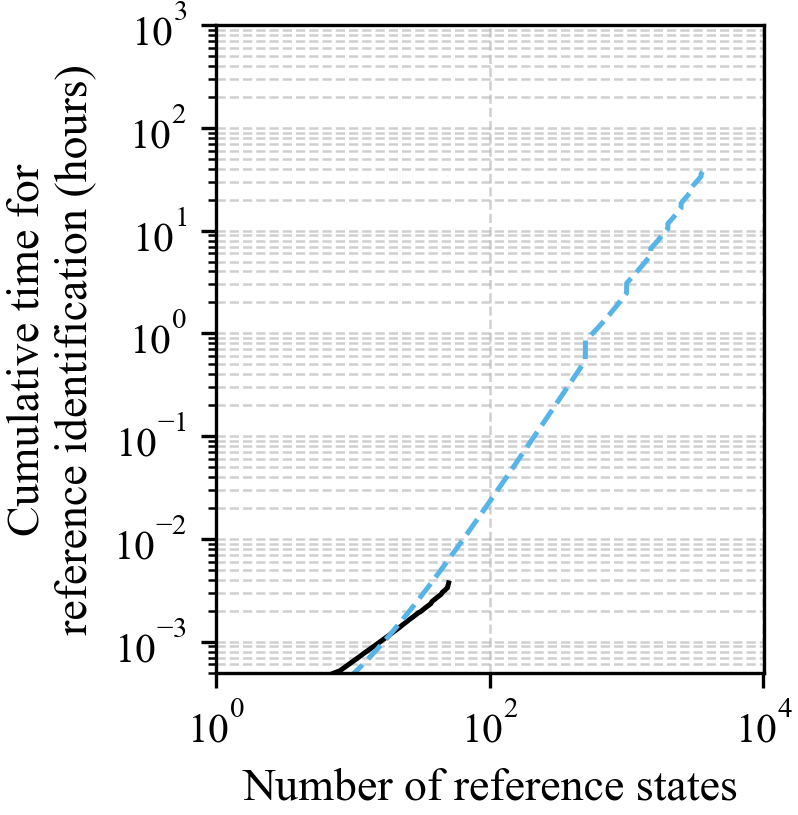}
        \caption{}
        \label{subfig:rsr_time_nomin}
    \end{subfigure}
    \hfill   
    \begin{subfigure}[t]{0.31\textwidth}
        \centering
        \includegraphics[width=\linewidth]{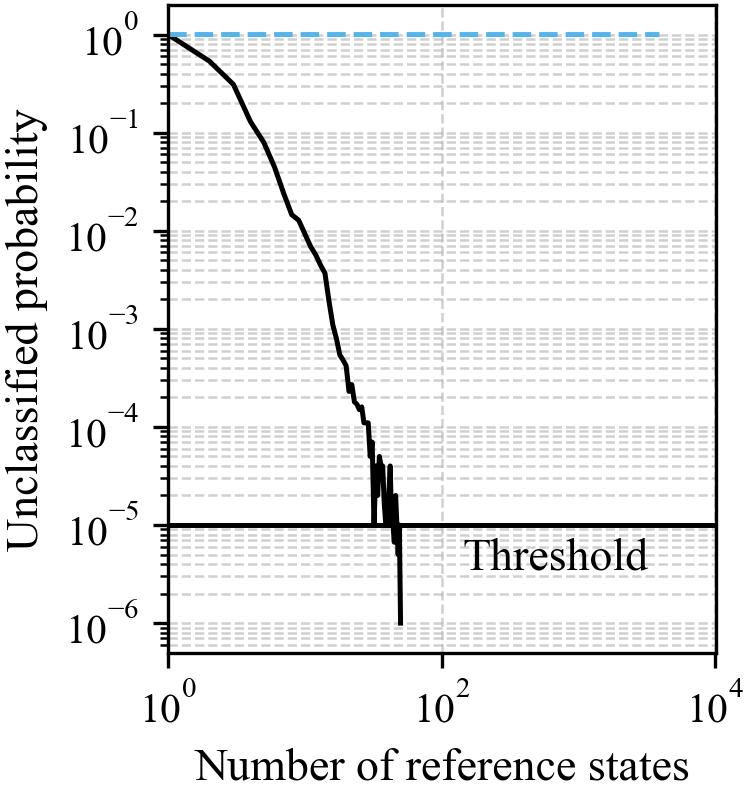}
        \caption{}
        \label{subfig:rsr_prob_nomin}
    \end{subfigure}
    
    \caption{\textbf{RSR implementation with and without the componentwise boundary search.} The results are shown for single-OD on the random graph in Figure~\ref{subfig:rg1}. Black solid lines and light blue dashed lines respectively correspond to with and without the componentwise boundary search. Black lines represent the identical data shown in Figures~\ref{subfig:brc_rsr_mem}--\ref{subfig:brc_rsr_prob}.
    (a) Memory usage measured by RSS, (b) cumulative computational time, and (c) unclassified probability, all shown as functions of the number of reference states.}
    \label{fig:rsr_res_nomin}
\end{figure}

We now examine the performance of the component-wise boundary search proposed in Section~\ref{subsec:new_ref}. Figure~\ref{fig:rsr_res_nomin} compares RSR's performance with and without this strategy for single-OD connectivity of Graph~1. As shown in Figures~\ref{subfig:rsr_mem_nomin} and~\ref{subfig:rsr_time_nomin}, memory usage and computation time follow similar trends in both cases as the number of reference states grows. The critical difference lies in Figure~\ref{subfig:rsr_prob_nomin}: without the componentwise boundary search, the unclassified probability remains close to 1.0 even after thousands of reference states are identified, indicating that almost no useful classification is achieved. With the search enabled, the unclassified probability reaches $10^{-5}$
with only 49 reference states. This demonstrates that the component-wise boundary search is essential to RSR's effectiveness, and that the additional system function evaluations it requires are well justified.

\subsection{Acceleration by high-performance computing}

The results in Section~\ref{subsec:rol_sys_per} indicate that solving the global connectivity problem for Graph~2 remains challenging on a desktop computer, with the unclassified probability stalling at $6.40\times10^{-2}$ after 10{,}000 reference states and 247~hours of computation. Since RSR's reference-state identification is inherently parallel---each candidate reference state can be evaluated independently---we now examine the gains achievable through high-performance computing (HPC). The following HPC computation is performed on NCI Gadi (gpuvolta queue) using two NVIDIA V100 GPUs (32~GB memory each), 24 CPU cores (Intel Xeon Platinum 8268), and 96 parallel workers.

Table~\ref{tab:rsr_hpc} summarises the HPC results for global connectivity, with percentage ratios reported relative to the desktop baselines in Table~\ref{tab:global_vs_singleod}. For Graph~1, the reference search time drops from 17.2~hours to 1.12~hours---a reduction of over 93\%---while reaching the same unclassified probability threshold of $1\times10^{-5}$ with slightly fewer reference states (6{,}757 vs.\ 8{,}680). Memory usage remains similar at 0.88~GB. On the other hand, Graph~2 shows more striking impact. With the reference-state limit raised to accommodate HPC capacity, RSR identifies over 560{,}000 reference states in under 5~hours, reducing the unclassified probability from $6.40\times10^{-2}$ to $2.22\times10^{-3}$---a factor of nearly 30. This transforms a problem that was practically intractable on a desktop into one that can be addressed within a few hours on a modest HPC allocation. However, the memory usage for Graph~2 rises to 9.73~GB, reflecting the cumulative storage of over half a million reference-state matrices. Moreover, even with this many reference states, the unclassified probability remains two orders of magnitude above the $10^{-5}$ target, implying a fundamental limitation of explicit reference-state enumeration as system complexity grows.

\begin{figure}[H]
    \centering
    \begin{subfigure}[t]{0.31\textwidth}
        \centering
        \includegraphics[width=\linewidth]{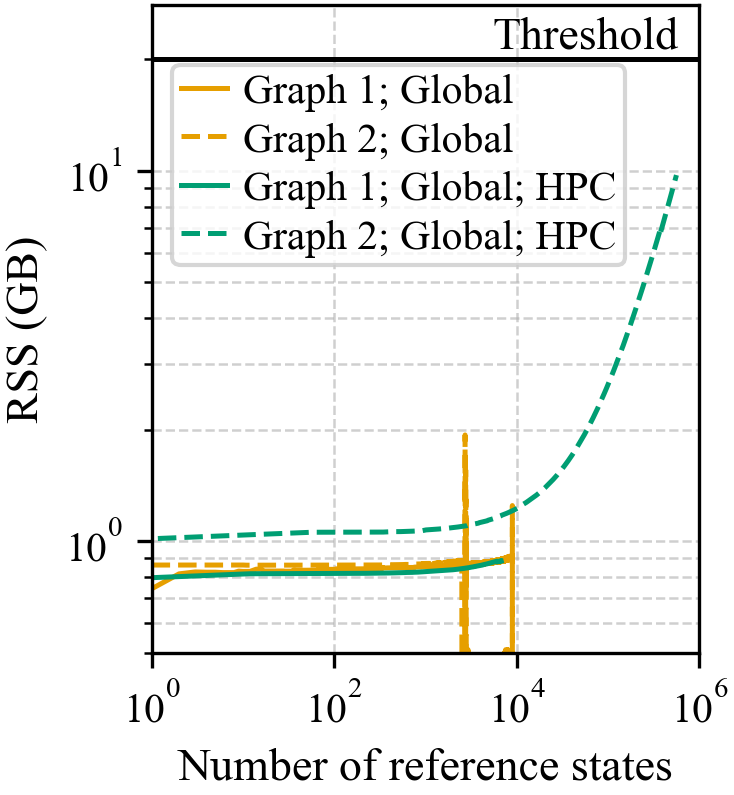}
        \caption{}
        \label{subfig:rsr_mem_hpc}
    \end{subfigure}
    \hfill
    \begin{subfigure}[t]{0.31\textwidth}
        \centering
        \includegraphics[width=\linewidth]{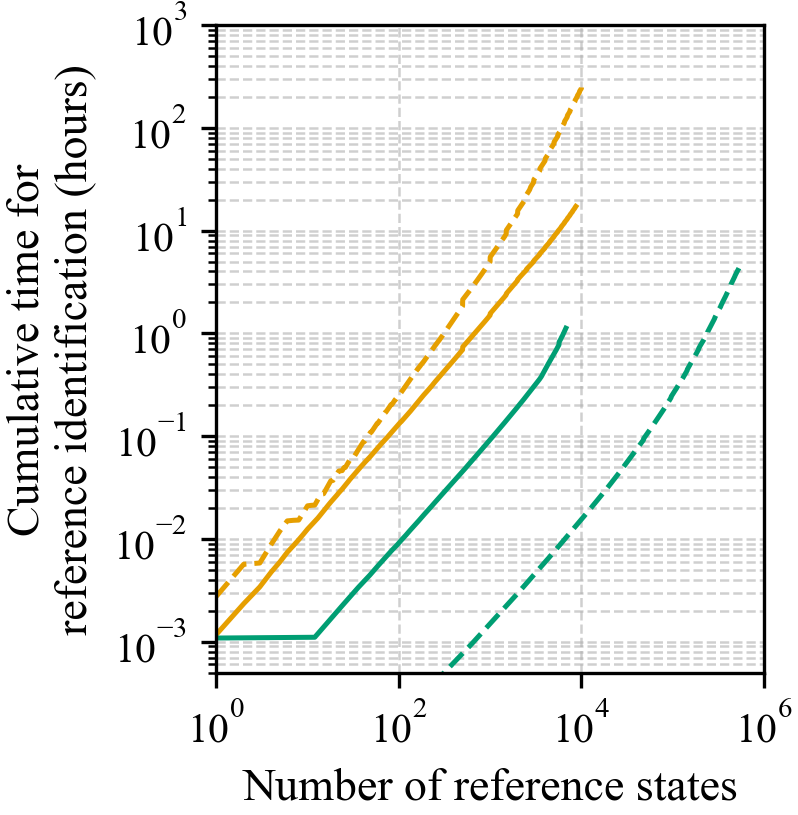}
        \caption{}
        \label{subfig:rsr_time_gpu}
    \end{subfigure}
    \hfill   
    \begin{subfigure}[t]{0.31\textwidth}
        \centering
        \includegraphics[width=\linewidth]{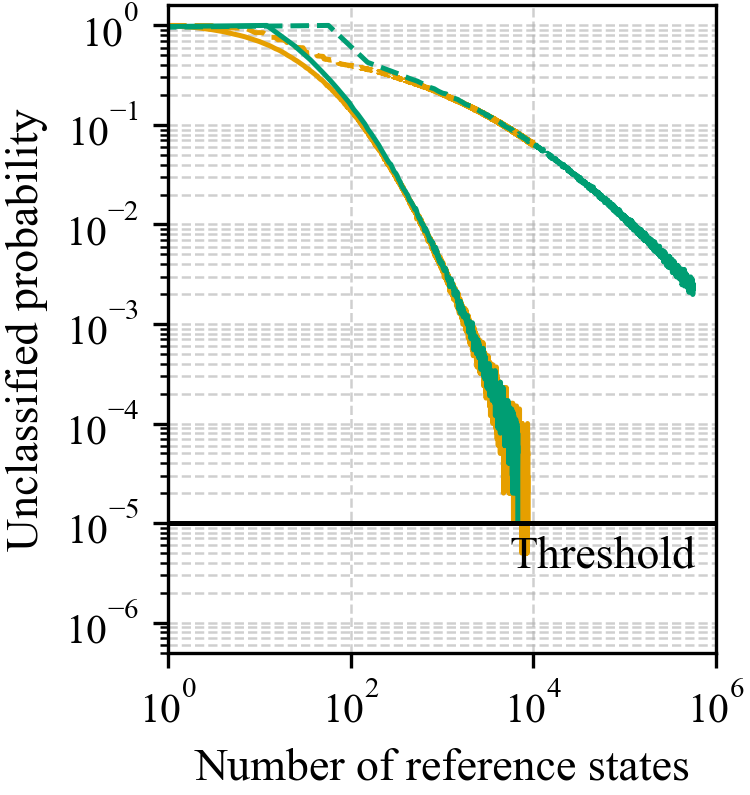}
        \caption{}
        \label{subfig:rsr_prob_gpu}
    \end{subfigure}
    
    \caption{\textbf{Comparison of computational performance by desktop and high-performance computing} on NCI Gadi (GPU queue, 2 GPUs, 24 CPU cores, 96 workers).
    (a) Memory usage measured by RSS, (b) cumulative computational time, and (c) unclassified probability, all shown as functions of the number of reference states.
    Black lines represent single-OD connectivity and are identical to the data shown in Figures~\ref{subfig:brc_rsr_mem}--\ref{subfig:brc_rsr_prob}, with the x-axes shown on a log scale to facilitate comparison.
    Orange lines represent global connectivity.
    Solid lines correspond to Graph~1 and dashed lines to Graph~2.}
    \label{fig:rsr_res_gpu}
\end{figure}

Figure~\ref{fig:rsr_res_gpu} provides further detail on the scaling behaviour under HPC. As shown in Figure~\ref{subfig:rsr_mem_hpc}, memory usage remains stable for most of the computation but begins to increase once the number of reference states grows sufficiently large, as the cumulative storage of reference-state matrices eventually dominates the baseline footprint. This is particularly evident for Graph~2, where memory rises sharply beyond $10^5$ reference states. While 9.73~GB remains well within practical limits, this trend suggests that memory may become a binding constraint for even larger problems. The cumulative search time (Figure~\ref{subfig:rsr_time_gpu}) increases at a nearly constant rate, consistent with the desktop results but at a dramatically lower per-iteration cost. Figure~\ref{subfig:rsr_prob_gpu} shows that the unclassified probability for Graph~2 continues to decrease steadily with additional reference states, suggesting that additional computation would yield further reductions, albeit at a diminishing rate.

\begin{table}[H]
    \centering
    \begin{tabular}{l|c c|c c}
    \Xhline{1.2pt}
        \multirow{2}{*}{} 
        & \multicolumn{2}{c|}{Graph 1}  
        & \multicolumn{2}{c}{Graph 2} \\
        \cline{2-5}
        & \makecell[c]{HPC}
        & \makecell[c]{HPC / Desktop}
        & \makecell[c]{HPC}
        & \makecell[c]{HPC /Desktop} \\
    \Xhline{1.2pt} 
         \makecell[l]{No. of reference states} 
         & 6{,}757 
         & $78 \ \%$ 
         & 560{,}597 & $5{,}606 \ \%$
         \\
         Memory (RSS, GB)
         & 0.882
         & 96 \%
         & 9.73
         & 2,812 \% \\
         \makecell[l]{Reference search\\[-5pt]time (hours)} 
         & 1.12 
         & $6.5 \ \%$ 
         & 4.77 & $1.9 \ \%$ \\
         \makecell[l]{Unclassified prob.} 
         & $1.00 \cdot 10^{-5}$ 
         & $100 \ \%$ 
         & $2.22 \cdot 10^{-3}$ & $3.5 \ \%$ \\
    \Xhline{1.2pt}
    \end{tabular}
    \caption{\textbf{RSR performance for global connectivity with HPC-enabled parallel multi-GPU execution on the random graphs in Figure~\ref{fig:brc}.} Percentage ratios are relative to the desktop results in Table~\ref{tab:global_vs_singleod}. The termination criteria are as described in Table~\ref{tab:global_vs_singleod}, with the reference-state limit raised to accommodate HPC capacity. HPC reduces the reference search time by over 93\% for Graph~1 and 98\% for Graph~2, and for Graph~2 enables identification of over 560{,}000 reference states, reducing the unclassified probability from $6.40 \cdot 10^{-2}$ to $2.22 \cdot 10^{-3}$.}
    \label{tab:rsr_hpc}
\end{table}

\subsection{Application to multi-state system}

The preceding analyses have considered binary system states. We now demonstrate RSR's applicability to multi-state systems by defining three performance levels for Graph~1 based on global connectivity: state~0 (loss of global connectivity), state~1 (global connectivity maintained with at most one edge-disjoint path between all OD pairs), and state~2 (global connectivity maintained with two or more edge-disjoint paths between all OD pairs). To handle multiple states, two separate analyses are performed with target system states $m'=0$ and $m'=1$, each run until the unclassified probability falls below $1\times10^{-5}$ for all boundaries.

Table~\ref{tab:rsr_multistate} summarises the results. The analysis with $m'=0$ reuses the results from Table~\ref{tab:global_vs_singleod}, while the analysis with $m'=1$ introduces a new boundary and requires 10{,}624 additional reference states. The system-state probabilities are obtained as $P(S=0) = P(S\le 0) = 2.47\times10^{-3}$, $P(S=1) = P(S\le 1) - P(S\le 0) = 1.02\times10^{-1}$, and $P(S=2) = 1 - P(S\le 1) = 8.96\times10^{-1}$, indicating that Graph~1 operates at the highest connectivity level with approximately 90\% probability.

\begin{table}[H]
    \centering
    \begin{tabular}{l|c c| c c}
    \Xhline{1.2pt}
        & $S \le 0$ & $S \ge 1$ & $S \le 1$ & $S \ge 2$ \\
    \Xhline{1.2pt}
         \makecell[l]{No. of reference states}
         & 19 & 8{,}661 & 150 & 27{,}735 \\
         \makecell[l]{Probability}
         & $2.47 \cdot 10^{-3}$
         & $9.97 \cdot 10^{-1}$ 
         & $1.04 \cdot 10^{-1}$ &
         $8.96 \cdot 10^{-1}$ \\
    \Xhline{1.2pt}
    \end{tabular}
    \caption{\textbf{RSR performance for a three-state system on Graph~1.} System states are defined by global connectivity level: state~0 (disconnected), state~1 (connected with single edge-disjoint paths), and state~2 (connected with two or more edge-disjoint paths). Each boundary is analysed until the unclassified probability falls below $1\times10^{-5}$.}
    \label{tab:rsr_multistate}
\end{table}

\section{Conclusions}\label{sec:con}

This study proposed the \textit{Reference-State System Reliability} (RSR) method for efficient uncertainty quantification of \textit{coherent} systems. Rather than decomposing the component-state space as in previous approaches, RSR generates Monte Carlo samples and classifies them into system states using \textit{reference states}---compact representations of the boundary between different system states. The method operates in two stages: Stage~1 iteratively identifies reference states forming these boundaries, while Stage~2 uses the identified reference states to classify Monte Carlo samples and estimate system probabilities. Once identified, reference states can be reused under changes in component probabilities, inheriting a key advantage of decomposition-based approaches.

RSR was benchmarked against the BRC algorithm \citep{BRC2026} for single-origin-destination connectivity of large-scale random graphs, achieving up to a 24-fold reduction in memory usage and more than a three-order-of-magnitude reduction in computation time. Classification of $10^6$ Monte Carlo samples required only 1.33~seconds and 9.33~seconds for graphs with 59 nodes/262 edges and 119 nodes/295 edges, respectively, demonstrating the significant improvement of computational cost. RSR was further applied to global connectivity of the same graphs---a problem beyond the reach of existing decomposition-based methods. The relatively uniform contribution of each component substantially increased the number of reference states, with RSR identifying 8{,}680 and over 560{,}000 reference states for the two graphs. Parallelisation via high-performance computing reduced the reference search time by over 93\% and 98\%, respectively, confirming that RSR's architecture is well suited to modern parallel computing environments. The method was also demonstrated on a three-state system, demonstrating its applicability to multi-state systems.

The design of RSR reflects a broader shift in scientific computing: rather than relying on a few powerful CPUs, modern workloads increasingly exploit the massive parallelism offered by GPUs. RSR embraces this paradigm by formulating both reference-state search and probability evaluation as operations over large batches of independent samples, enabling full parallelisation across workers and GPU cores. Advanced sampling techniques such as importance sampling could, in principle, reduce the number of samples required---but at the cost of introducing inter-sample dependencies that would break this parallelism. The resulting slowdown per iteration may offset any reduction in the total number of iterations. While further investigation is needed, this trade-off suggests a potentially fruitful direction for system reliability analysis more broadly: designing methods around parallelisable computations rather than sample-efficient but inherently sequential ones.

Despite RSR's substantially improved scalability, the global connectivity problem for the larger graph reveals an inherent limitation shared by RSR and other decomposition-based methods: the explicit enumeration of reference states. Even after identifying over 560{,}000 reference states, the unclassified probability for Graph~2 remains at $2.22\times10^{-3}$, with further identification yielding diminishing returns. This suggests that highly complex systems may possess a prohibitively large number of influential reference states. This observation motivates further research into more efficient representations of reference-state boundaries---for example, machine-learning approaches that exploit structural patterns among previously identified reference states to generalise without exhaustive enumeration.

Overall, RSR demonstrates that coherent systems with hundreds of components can be analysed at second- to minute-scale, depending on the complexity of the system performance function. Incoherent systems, by contrast, remain largely an open challenge. In practice, many real-world systems exhibit coherency under normal operating conditions, with local incoherency arising only in a small proportion of extreme scenarios---for example, when collectively suboptimal decisions are made by users or operators. Such systems could potentially be addressed in future work through reference-state search strategies that exploit \textit{weak} coherency, building on the framework developed in this study.

\section*{Acknowledgments}

\noindent  This research was undertaken with the assistance of resources from the National Computational Infrastructure (NCI Australia), an NCRIS enabled capability supported by the Australian Government. The corresponding author was supported by the National Research Foundation of Korea (NRF) grant funded by the Korea government (MSIT) (RS-2026-25470577) and the Institute of Construction and Environmental Engineering at Seoul National University.

\section*{Data availability}
\noindent RSR is available as a Python package at \url{https://github.com/jieunbyun/rsr/}, along with example data and scripts.

\bibliographystyle{elsarticle-num-names} 
\bibliography{refs}

@article{WuSun2024_layeredRD,
  author  = {Wu, Baichao and Sun, Long},
  title   = {A novel layer-by-layer recursive decomposition algorithm for calculation of network reliability},
  journal = {Reliability Engineering \& System Safety},
  volume  = {244},
  pages   = {109968},
  year    = {2024}
}

@article{LiHe2002_recursive,
  author  = {Li, Jie and He, Jun},
  title   = {A recursive decomposition algorithm for network seismic reliability evaluation},
  journal = {Earthquake Engineering \& Structural Dynamics},
  volume  = {31},
  number  = {8},
  pages   = {1525--1539},
  year    = {2002}
}

@article{DalyAlexopoulos2006_statespace,
  author  = {Daly, Matthew S. and Alexopoulos, Christos},
  title   = {State-space partition techniques for multiterminal flows in stochastic networks},
  journal = {Networks: An International Journal},
  volume  = {48},
  number  = {2},
  pages   = {90--111},
  year    = {2006}
}

@article{LimSong2012_lifeline,
  author  = {Lim, Hyun-Woo and Song, Junho},
  title   = {Efficient risk assessment of lifeline networks under spatially correlated ground motions using selective recursive decomposition algorithm},
  journal = {Earthquake Engineering \& Structural Dynamics},
  volume  = {41},
  number  = {13},
  pages   = {1861--1882},
  year    = {2012}
}

@article{JanLai08,
    author = "Jane, C. C. and Laih, Y. W.",
    title = "A practical algorithm for computing multi-state two-terminal reliability",
    journal = "IEEE Transactions on reliability",
    year = "2008",
    volume = "57",
    number = "2",
    pages = "295--302"
}

@article{Chang2024_pathbased, 
  author  = {Chang, Ping-Chen},
  title   = {A path-based simulation approach for multistate flow network reliability estimation without using boundary points},
  journal = {Reliability Engineering \& System Safety},
  volume  = {249},
  pages   = {110237},
  year    = {2024}
}

@article{Niu2025_multicommodity,
  author  = {Niu, Yi-Feng and Wang, Jun-Feng and Xu, Xiu-Zhen and Xu, Qian-Xin},
  title   = {Reliability evaluation for a multi-commodity multi-state distribution network under transportation emission consideration},
  journal = {Reliability Engineering \& System Safety},
  volume  = {254},
  pages   = {110599},
  year    = {2025}
}

@article{Yeh2021_BAT,
  author  = {Yeh, Wei-Chang},
  title   = {Novel binary-addition tree algorithm (BAT) for binary-state network reliability problem},
  journal = {Reliability Engineering \& System Safety},
  volume  = {208},
  pages   = {107448},
  year    = {2021}
}

@article{Yeh2022_SA-BAT,
  author  = {Yeh, Wei-Chang},
  title   = {Novel self-adaptive Monte Carlo simulation based on binary-addition-tree algorithm for binary-state network reliability approximation},
  journal = {Reliability Engineering \& System Safety},
  volume  = {228},
  pages   = {108796},
  year    = {2022}
}

@article{Zhang2025_multistatecuts,
  author  = {Zhang, Shuai and Bai, Guanghan and Tao, Junyong and Wang, Yang and Xu, Bei},
  title   = {An algorithm to search for multi-state minimal cuts in multi-state flow networks containing state heterogeneous components},
  journal = {Reliability Engineering \& System Safety},
  volume  = {256},
  pages   = {110774},
  year    = {2025}
}

@article{Kozyra2025_simulation,
  author  = {Kozyra, Pawe{\l} Marcin},
  title   = {A simulation approach with heuristic rules for reliability estimation of two-terminal multi-state networks based on minimal cuts and parallel computations},
  journal = {Simulation Modelling Practice and Theory},
  volume  = {141},
  pages   = {103095},
  year    = {2025}
}

@article{Chang2025_logistics,
  author  = {Chang, Ping-Chen and Yeh, Cheng-Ta and Cheng, Tzu-Yun and Yu, Vincent F.},
  title   = {Reliability Assessment for Multistate Semiconductor Global Logistics Networks with Freight Charges and Carbon Taxes Considerations},
  journal = {Reliability Engineering \& System Safety},
  year    = {2025},
  pages   = {112027}
}

@article{Huang2024_LP_MC,
  author  = {Huang, Ding-Hsiang},
  title   = {Network reliability of a stochastic flow network by wrapping linear programming models into a Monte-Carlo simulation},
  journal = {Reliability Engineering \& System Safety},
  volume  = {252},
  pages   = {110427},
  year    = {2024}
}

@article{ChenLin2012_minpaths,
  author  = {Chen, Shin-Guang and Lin, Yi-Kuei},
  title   = {Search for all minimal paths in a general large flow network},
  journal = {IEEE Transactions on Reliability},
  volume  = {61},
  number  = {4},
  pages   = {949--956},
  year    = {2012}
}

@article{Liu2025_configuration,
  author  = {Liu, Tao and Liu, Jun and Bai, Guanghan and Zhang, Junfu and Wang, Libo},
  title   = {A reliability evaluation method for multi-state flow networks considering network configuration adjustment},
  journal = {Reliability Engineering \& System Safety},
  volume  = {261},
  pages   = {111158},
  year    = {2025}
}

@article{Xu2025_dminimal,
  author  = {Xu, Bei and Hu, Yi-fan and Zhang, Huang-xiao and Zhu, Yan-ping and Bai, Guang-han},
  title   = {Improved algorithm for d-minimal cuts in reliability evaluation of multi-state flow networks},
  journal = {Reliability Engineering \& System Safety},
  year    = {2025},
  pages   = {111707}
}

@article{Zuo2007_minpathvectors,
  author  = {Zuo, Ming J. and Tian, Zhigang and Huang, Hong-Zhong},
  title   = {An efficient method for reliability evaluation of multistate networks given all minimal path vectors},
  journal = {IIE Transactions},
  volume  = {39},
  number  = {8},
  pages   = {811--817},
  year    = {2007}
}

@article{Bai2018_statespace,
  author  = {Bai, Guanghan and Tian, Zhigang and Zuo, Ming J.},
  title   = {Reliability evaluation of multistate networks: An improved algorithm using state-space decomposition and experimental comparison},
  journal = {IISE Transactions},
  volume  = {50},
  number  = {5},
  pages   = {407--418},
  year    = {2018}
}

@article{Chang2022_simulation,
  author  = {Chang, Ping-Chen},
  title   = {Simulation approaches for multi-state network reliability estimation: Practical applications},
  journal = {Simulation Modelling Practice and Theory},
  volume  = {115},
  pages   = {102457},
  year    = {2022}
}

@article{LinHuang2014_multisink_accuracy,
  author  = {Lin, Yi-Kuei and Huang, Cheng-Fu},
  title   = {Reliability evaluation of a multi-state network with multiple sinks under individual accuracy rate constraint},
  journal = {Communications in Statistics -- Theory and Methods},
  volume  = {43},
  number  = {21},
  pages   = {4519--4533},
  year    = {2014}
}

@article{Zaitseva2025_BDD_uncertainty,
  author  = {Zaitseva, Elena and Levashenko, Vitaly},
  title   = {Reliability Analysis Based on Aleatory and Epistemic Uncertainty Using Binary Decision Diagrams},
  journal = {International Journal of Intelligent Systems},
  volume  = {2025},
  number  = {1},
  pages   = {6471577},
  year    = {2025}
}

@article{Zaitseva2023_MDD_epistemic,
  author  = {Zaitseva, Elena and Levashenko, Vitaly and Rabcan, Jan},
  title   = {A new method for analysis of multi-state systems based on multi-valued decision diagram under epistemic uncertainty},
  journal = {Reliability Engineering \& System Safety},
  volume  = {229},
  pages   = {108868},
  year    = {2023}
}

@article{Mo2013_BDD_ordering,
  author  = {Mo, Yuchang and Zhong, Farong and Liu, Huawen and Yang, Quansheng and Cui, Gang},
  title   = {Efficient ordering heuristics in binary decision diagram--based fault tree analysis},
  journal = {Quality and Reliability Engineering International},
  volume  = {29},
  number  = {3},
  pages   = {307--315},
  year    = {2013}
}

@article{Shrestha2009_MSS_DD,
  author  = {Shrestha, Akhilesh and Xing, Liudong and Dai, Yuanshun},
  title   = {Decision diagram based methods and complexity analysis for multi-state systems},
  journal = {IEEE Transactions on Reliability},
  volume  = {59},
  number  = {1},
  pages   = {145--161},
  year    = {2009}
}

@article{Xing2007_BDD,
  author  = {Xing, Liudong},
  title   = {An efficient binary-decision-diagram-based approach for network reliability and sensitivity analysis},
  journal = {IEEE Transactions on Systems, Man, and Cybernetics -- Part A: Systems and Humans},
  volume  = {38},
  number  = {1},
  pages   = {105--115},
  year    = {2007}
}

@article{Imai1999_allterminalBDD,
  author  = {Imai, Hiroshi and Sekine, Kyoko and Imai, Keiko},
  title   = {Computational investigations of all-terminal network reliability via BDDs},
  journal = {IEICE Transactions on Fundamentals of Electronics, Communications and Computer Sciences},
  volume  = {82},
  number  = {5},
  pages   = {714--721},
  year    = {1999}
}

@article{Niu2025_resourceconstrained,
  author  = {Niu, Yi-Feng and Yan, Yi-Fan and Xu, Xiu-Zhen},
  title   = {A new MC-based method for the resource-constrained multi-distribution multi-state flow network reliability optimization problem},
  journal = {Reliability Engineering \& System Safety},
  year    = {2025},
  pages   = {111499}
}

@article{Dong2016_DD_multistateflow,
  author  = {Dong, Rongsheng and Zhu, Yangyang and Xu, Zhoubo and Li, Fengying},
  title   = {Decision diagram based symbolic algorithm for evaluating the reliability of a multistate flow network},
  journal = {Mathematical Problems in Engineering},
  volume  = {2016},
  number  = {1},
  pages   = {6908120},
  year    = {2016}
}

@article{BRC2026,
    author = {Byun, Ji-Eun and Ryu, Hyeuk and Straub, Daniel},
    title = {Branch-and-bound algorithm for efficient reliability analysis of general coherent systems},
    journal = {Structural Safety},
    year = {2026},
    volume = {118},
    pages = {102653}
}

@article{Zhang2019_MDD_trust,
  author  = {Zhang, Li and Xing, Liudong and Liu, Anqing and Mao, Keming},
  title   = {Multivalued decision diagrams-based trust level analysis for social networks},
  journal = {IEEE Access},
  volume  = {7},
  pages   = {180620--180629},
  year    = {2019}
}

@article{Song2009_matrix_dependence,
  author  = {Song, Junho and Kang, Won-Hee},
  title   = {System reliability and sensitivity under statistical dependence by matrix-based system reliability method},
  journal = {Structural Safety},
  volume  = {31},
  number  = {2},
  pages   = {148--156},
  year    = {2009}
}

@article{Xing2025_decisiondiagrams,
  author  = {Xing, L.},
  title   = {A review of decision diagrams in system reliability modeling and analysis},
  journal = {Applied Mathematical Modelling},
  year    = {2025},
  pages   = {116039}
}

@article{Jiang2025_hierarchical,
  author  = {Jiang, Chen and Sun, Muxia and Wang, Luyao and Wang, Zisheng and Li, Yan-Fu},
  title   = {A recursive algorithm for reliability evaluation of multi-state hierarchical systems with stochastic dependent components},
  journal = {Reliability Engineering \& System Safety},
  year    = {2025},
  pages   = {111653}
}

@article{Lee2025_DGBN_RB,
  author  = {Lee, Dongkyu and Byun, Ji-Eun and Song, Junho},
  title   = {Dual graph-based {B}ayesian network modeling with {R}ao-{B}lackwellization for seismic reliability and complexity quantification of network connectivity},
  journal = {Earthquake Engineering \& Structural Dynamics},
  year    = {2025},
  volume = {54},
  number = {10},
  pages = {2387--2402}
}

@article{LeeSong2021_multiscale,
  author  = {Lee, Dongkyu and Song, Junho},
  title   = {Multi-scale seismic reliability assessment of networks by centrality-based selective recursive decomposition algorithm},
  journal = {Earthquake Engineering \& Structural Dynamics},
  volume  = {50},
  number  = {8},
  pages   = {2174--2194},
  year    = {2021}
}

@article{Herrera2025_landmark,
  author  = {Herrera, Manuel and Giudicianni, Carlo and Sasidharan, Manu and Wright, Robert and Creaco, Enrico and Parlikad, Ajith Kumar},
  title   = {Landmark-node based reliability assessment for critical infrastructure networks},
  journal = {Reliability Engineering \& System Safety},
  year    = {2025},
  pages   = {111563}
}

@article{Li2023_bridgeconnectivity,
  author  = {Li, Shunlong and Wang, Jie and He, Shaoyang},
  title   = {Connectivity probability evaluation of a large-scale highway bridge network using network decomposition},
  journal = {Reliability Engineering \& System Safety},
  volume  = {236},
  pages   = {109191},
  year    = {2023}
}

@article{Miao2020_PDEM_water,
  author  = {Miao, Huiquan and Liu, Wei and Li, Jie},
  title   = {Seismic reliability analysis of water distribution networks on the basis of the probability density evolution method},
  journal = {Structural Safety},
  volume  = {86},
  pages   = {101960},
  year    = {2020}
}

@article{Liu2018_PDEM_gas,
  author  = {Liu, Wei and Li, Zongcai and Song, Zhaoyang and Li, Jie},
  title   = {Seismic reliability evaluation of gas supply networks based on the probability density evolution method},
  journal = {Structural Safety},
  volume  = {70},
  pages   = {21--34},
  year    = {2018}
}

@article{Shi2024_survivalsignatureML,
  author  = {Shi, Yan and Behrensdorf, Jasper and Zhou, Jiayan and Hu, Yue and Broggi, Matteo and Beer, Michael},
  title   = {Network reliability analysis through survival signature and machine learning techniques},
  journal = {Reliability Engineering \& System Safety},
  volume  = {242},
  pages   = {109806},
  year    = {2024}
}

@article{HuaHuaLin22,
  title={A novel approach to predict network reliability for multistate networks by a deep neural network},
  author={Huang, C. H. and Huang, D. H. and Lin, Y. K.},
  journal={Quality Technology \& Quantitative Management},
  volume={19},
  number={3},
  pages={362--378},
  year={2022}
}

@article{HuaHuaLin23,
  title={Network reliability prediction for random capacitated-flow networks via an artificial neural network},
  author={Huang, Chun Hsiang and Huang, Da Hsin and Lin, Yung Kun},
  journal={Reliability Engineering \& System Safety},
  volume={237},
  pages={109378},
  year={2023}
}

@article{HuangLin2024_LSTM,
  author  = {Huang, Cheng-Hao and Lin, Yi-Kuei},
  title   = {Manufacturing system evaluation in terms of system reliability via long short-term memory},
  journal = {Reliability Engineering \& System Safety},
  volume  = {251},
  pages   = {110365},
  year    = {2024}
}

@article{ByunSong2021_GMBN,
  author  = {Byun, Ji-Eun and Song, Junho},
  title   = {Generalized matrix-based {B}ayesian network for multi-state systems},
  journal = {Reliability Engineering \& System Safety},
  volume  = {211},
  pages   = {107468},
  year    = {2021}
}

@article{DuenasOsorio2018_quantuminspired,
  author  = {Due{\~n}as-Osorio, Leonardo and Vardi, Moshe and Rojo, Javier},
  title   = {Quantum-inspired boolean states for bounding engineering network reliability assessment},
  journal = {Structural Safety},
  volume  = {75},
  pages   = {110--118},
  year    = {2018}
}

@article{Zuev2015_subset,
  author  = {Zuev, Konstantin M. and Wu, Stephen and Beck, James L.},
  title   = {General network reliability problem and its efficient solution by subset simulation},
  journal = {Probabilistic Engineering Mechanics},
  volume  = {40},
  pages   = {25--35},
  year    = {2015}
}

@article{Vaisman2016_SSMC,
  author  = {Vaisman, R. and Kroese, D. P. and Gertsbakh, I. B.},
  title   = {Splitting sequential Monte Carlo for efficient unreliability estimation of highly reliable networks},
  journal = {Structural Safety},
  volume  = {63},
  pages   = {1--10},
  year    = {2016}
}

@article{Chan2024_BICE,
  author  = {Chan, Jianpeng and Papaioannou, Iason and Straub, Daniel},
  title   = {Bayesian improved cross entropy method with categorical mixture models for network reliability assessment},
  journal = {Reliability Engineering \& System Safety},
  volume  = {252},
  pages   = {110432},
  year    = {2024}
}

@article{Chan2023_BICE,
  author  = {Chan, Jianpeng and Papaioannou, Iason and Straub, Daniel},
  title   = {Bayesian improved cross entropy method for network reliability assessment},
  journal = {Structural Safety},
  volume  = {103},
  pages   = {102344},
  year    = {2023}
}

@article{Kanjilal2023_CEIS,
  author  = {Kanjilal, Oindrila and Papaioannou, Ioannis and Straub, Daniel},
  title   = {Bayesian updating of reliability by cross entropy-based importance sampling},
  journal = {Structural Safety},
  volume  = {102},
  pages   = {102325},
  year    = {2023}
}

@article{Lee2025_subset_lifeline,
  author  = {Lee, Dongkyu and Wang, Ziqi and Song, Junho},
  title   = {Efficient seismic reliability and fragility analysis of lifeline networks using subset simulation},
  journal = {Reliability Engineering \& System Safety},
  volume  = {260},
  pages   = {110947},
  year    = {2025}
}

@inproceedings{ByunStraub2023_ICASP,
  author    = {Byun, Ji-Eun and Straub, Daniel},
  title     = {Efficient reliability analysis of coherent systems with discrete-state components},
  booktitle = {Proceedings of the 14th International Conference on Applications of Statistics and Probability in Civil Engineering (ICASP14)},
  year      = {2023},
  address   = {Dublin, Ireland},
}

@article{Zwirglmaier2024_HybridBN,
  author  = {Zwirglmaier, Kilian and Chan, Jianpeng and Papaioannou, Iason and Song, Junho and Straub, Daniel},
  title   = {Hybrid {B}ayesian Networks for Reliability Assessment of Infrastructure Systems},
  journal = {ASCE-ASME Journal of Risk and Uncertainty in Engineering Systems, Part A: Civil Engineering},
  volume  = {10},
  number  = {2},
  year    = {2024},
  pages   = {04024019},
}

\end{document}